\documentclass[10pt,twocolumn,letterpaper]{article}

\usepackage{3dv}
\usepackage{times}
\usepackage{epsfig}
\usepackage{amsmath}
\usepackage{amssymb}
\usepackage{subfigure}
\usepackage{booktabs}
\usepackage{float}
\usepackage[bottom]{footmisc}

\DeclareFontFamily{U}{dutchcal}{\skewchar\font=45 }
\DeclareFontShape{U}{dutchcal}{m}{n}{<-> s*[1.0] dutchcal-r}{}
\DeclareFontShape{U}{dutchcal}{b}{n}{<-> s*[1.0] dutchcal-b}{}
\DeclareMathAlphabet{\mathlcal}{U}{dutchcal}{m}{n}
\SetMathAlphabet{\mathlcal}{bold}{U}{dutchcal}{b}{n}

\newcommand{\argmin}[1]{\underset{#1}{\operatorname{arg}\,\operatorname{min}}\;}

\newcommand{\startcompact}[1]{\par\vspace{-0.75em}\begin{#1}%
\allowdisplaybreaks\ignorespaces}
\newcommand{\stopcompact}[1]{\end{#1}\ignorespaces}

\newenvironment{packed_item}{
\begin{itemize}
   \setlength{\itemsep}{1pt}
   \setlength{\parskip}{0pt}
   \setlength{\parsep}{0pt}
 }{\end{itemize}}

\usepackage[pagebackref=true,breaklinks=true,letterpaper=true,colorlinks,bookmarks=false]{hyperref}

\threedvfinalcopy 

\def\threedvPaperID{13} 

\begin{document}
\title{Compact Model Representation for 3D Reconstruction}

\author{Jhony K. Pontes$^{\star \dagger}$, Chen Kong$^\star$, Anders Eriksson$^\dagger$, Clinton Fookes$^\dagger$ Sridha Sridharan$^\dagger$, and Simon Lucey$^\star$ \\
\begin{tabular}[h]{cc}
	Carnegie Mellon University$^\star$ & Queensland University of Technology $^\dagger$  \\
	{\tt\small \{jpontes, chenk, slucey\}@andrew.cmu.edu} &  {\tt\small \{anders.eriksson, c.fookes, s.sridharan\}@qut.edu.au}
  \end{tabular}      
}

\twocolumn[{%
\renewcommand\twocolumn[1][]{#1}%

\maketitle

\vspace{-1cm}
\begin{figure}[H]
\setlength{\hsize}{\textwidth}
\centering
\includegraphics[width=1.01\textwidth,keepaspectratio]{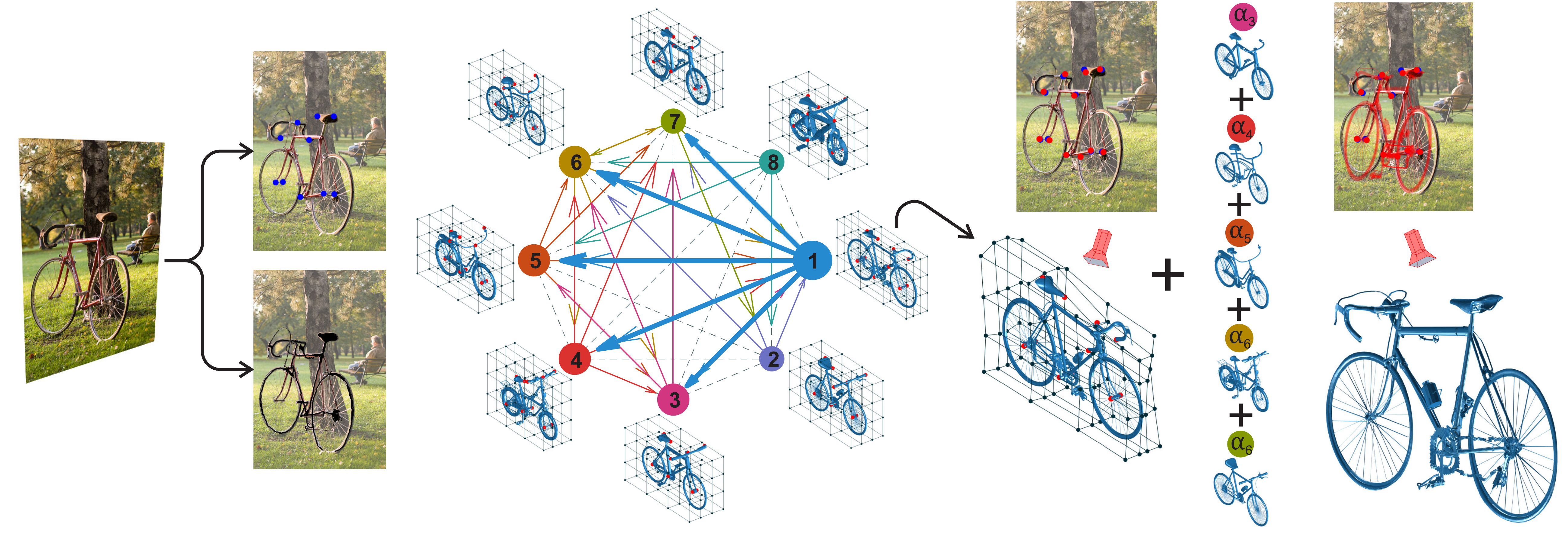} 
\caption{Given a 2D image, a few anchors and the silhouette, our method first selects a 3D model from a compact graph that best fits the 2D anchors. Second, the 3D model is refined by finding a linear combination (\ie $\alpha$'s) that best matches the image silhouette, \eg node 1 is the selected one so we can perform linear combination with the nodes 3, 4, 5, 6, and 7 (blue arrows).}
\label{fig:overview}
\end{figure}%
}]


\begin{abstract}
3D reconstruction from 2D images is a central problem in computer vision. Recent works have been focusing on reconstruction directly from a single image. It is well known however that only one image cannot provide enough information for such a reconstruction. A prior knowledge that has been entertained are 3D CAD models due to its online ubiquity. A fundamental question is how to compactly represent millions of CAD models while allowing generalization to new unseen objects with fine-scaled geometry. We introduce an approach to compactly represent a 3D mesh. Our method first selects a 3D model from a graph structure by using a novel free-form deformation (\textsc{Ffd}) 3D-2D registration, and then the selected 3D model is refined to best fit the image silhouette. We perform a comprehensive quantitative and qualitative analysis that demonstrates impressive dense and realistic 3D reconstruction from single images.
\end{abstract}

\section{Introduction}
A fundamental goal in computer vision is to infer from optical images to objects in the 3D world that give rise to them. This is often called the \textit{inverse problem} \cite{Palmer1999}. An obvious solution is to directly invert the optical transformations that occur during image formation. Unfortunately, this is an ``ill-defined" problem. Specifically, the inverse mapping from 2D to 3D is not unique. 

Vision researchers have recently started to consider using 3D prior knowledge to make the problem of 3D reconstruction from a single image less ambiguous \cite{Wang2014, Zhou2015, Zhou2016, Wu2016, Chen2016, Bansal2016, Han2016, Chen2017}.
3D reconstructions provided by these prior works are, however, either an object skeleton or a messy, unrealistic surface.
To handle this, deep learning techniques have been investigated to tackle this problem \cite{choy2016,Sharma2016,Maxim2017,Riegler2017}. It replaces the 2D pixel array by a dense and regular 3D voxel grid to process 3D data using 3D convolution and pooling operations.
This implies that the computational and memory requirements scale cubically with the resolution \cite{Maxim2017}, restricting volumetric representation to low 3D resolution and ``legolized", unrealistic 3D objects. 

In this work we exploit the rich and detailed geometry of real 3D objects. We propose a graph embedding to compactly represent 3D mesh objects that allow us to obtain reconstructions with much higher resolution. We tackle this challenge by revisiting the free-form deformation (\textsc{Ffd}) technique \cite{Sederberg1986}. \textsc{Ffd} is a classical, powerful and widely used graphics technique for deforming a 3D mesh \cite{Raffin2013}. To compactly model the intrinsic variation across a class of 3D objects, we use a low-dimensional \textsc{Ffd} parametrization and sparse linear representation in a dictionary. Our method first selects a 3D model from a graph structure by using a \textsc{Ffd} 3D-2D registration based on anchors, and then refines the selected 3D model by finding a sparse linear combination from the graph to fit the silhouette as shown in Figure~\ref{fig:overview}.



\noindent\textbf{Our core contributions are:}
\begin{packed_item}
\item We propose to compactly represent 3D mesh models by using only a few parameters - the displacements of the \textsc{Ffd} control points and the weights of the sparse linear combination;
\item We develop a \textsc{Ffd} 3D-2D registration algorithm to simultaneously select a 3D model from the graph that best fits an image and estimate the displacements of the \textsc{Ffd} control points based on its anchors;
\item Finally, we show empirically the utility of our framework for dense and more realistic 3D reconstructions from a single image.
\end{packed_item}

\subsection{Related Work}

In this subsection we review recent progress on 3D reconstruction and representation. Kong \etal \cite{Kong2016, Chen2016} recently proposed a method for learning a dictionary of 3D models from an ensemble of 2D anchors of a specific object category. Despite their work being able of handling highly deformable objects, the non-convex nature of the group-sparse dictionary learning proposed is sensitive to initialization and noise. Zhou \etal \cite{Zhou2015} learned a dictionary from a 3D shape dataset and proposed a convex relaxation technique to estimate the camera pose and the 3D shape parameters simultaneously given a single image and anchors. Deep learning techniques have also been entertained. Wu \etal \cite{Wu2016} trained a deep network to infer 3D points from 2D anchors. These works have achieved promising results, however they are only able to reconstruct 3D skeletons, non-detailed, and non-realistic 3D models mainly due to the difficulty of establishing dense correspondences between CAD models of the same class.  

Vicente \etal \cite{Vicente2014, Carreira2015} ambitiously obtained dense 3D reconstructions of rigid objects from single images. They used a set of images depicting different instances of the same object class to learn a ``generic'' class-based 3D model. A fundamental drawback of their approach is that the inferred 3D reconstruction is rigid. Kar \etal \cite{Kar2015} handled such a drawback by employing a novel dense surface model based on active shape models (ASM) to estimate a deformable dense hull of a single image. However, the approach is limited as the process smooths over important fine details in the reconstructed 3D object.

This paper builds upon the recent work of Kong \etal \cite{Chen2017} who proposed a novel graph embedding based on the local dense correspondences between 3D models. They showed that it is possible to perform sparse linear combination to deform a 3D model by using a dense shape dictionary. Their method is able to reconstruct the geometry of a single image given the anchors and silhouette. Despite their impressive results for dense 3D reconstruction, it still lacks of fine details and realism.

Another family of approaches related to our work are the ones using volumetric representations learned through deep neural networks \cite{Ulusoy2015,Cherabier2016,choy2016,Rezende2016,Yan2016,Sharma2016,Maxim2017,Riegler2017,Hane2017}. Of special interest is the recent work of Tatarchenko \etal \cite{Maxim2017} where they presented a deep convolutional decoder architecture to generate volumetric 3D outputs by using an octree representation which allows for much higher resolution outputs. Even though their method achieves high resolution and compact volumetric representations, their models lack fine-scaled geometry. A major advantage of our proposed method is that we have a compact representation of 3D models while keeping the finer geometry and object realism by using 3D meshes.

\subsection{Approach Overview}
Given a single 2D image of an object (\eg airplane, chair, car, etc.), our method estimates the camera pose and the 3D mesh model using 2D anchors and silhouette information. We assume we have the anchors and silhouette beforehand. We compactly represent a class of 3D models through a graph structure that allow us to perform 3D deformations by \textsc{Ffd} and linear combination. Once we build up the graph we can perform 3D reconstruction from a single image by first selecting a 3D model from the graph that best matches the image. After that, we refine the 3D model based on the image silhouette and the dense correspondences. Figure~\ref{fig:overview} shows an overview of the proposed method.    


\section{Compact Model Representation}
Our compact representation of 3D mesh models involves embedding all 3D models of a specific class into the \textsc{Ffd} space. By allowing \textsc{Ffd} for every model we can establish dense correspondences using nonrigid iterative closest point (ICP) \cite{Amberg2007}. Once we find dense correspondences for every pair in the class we build up a direct graph $\mathcal{G}(\mathcal{V}, \mathcal{E})$ where $\mathcal{V}$ are the vertices of the 3D models and $\mathcal{E}$ are the edges representing the deformed models with dense correspondences. 
\subsection{Free-Form Deformation}
\begin{figure}
	\centering
    \subfigure[]{
       	\includegraphics[width=100cm,height=2.8cm,keepaspectratio]{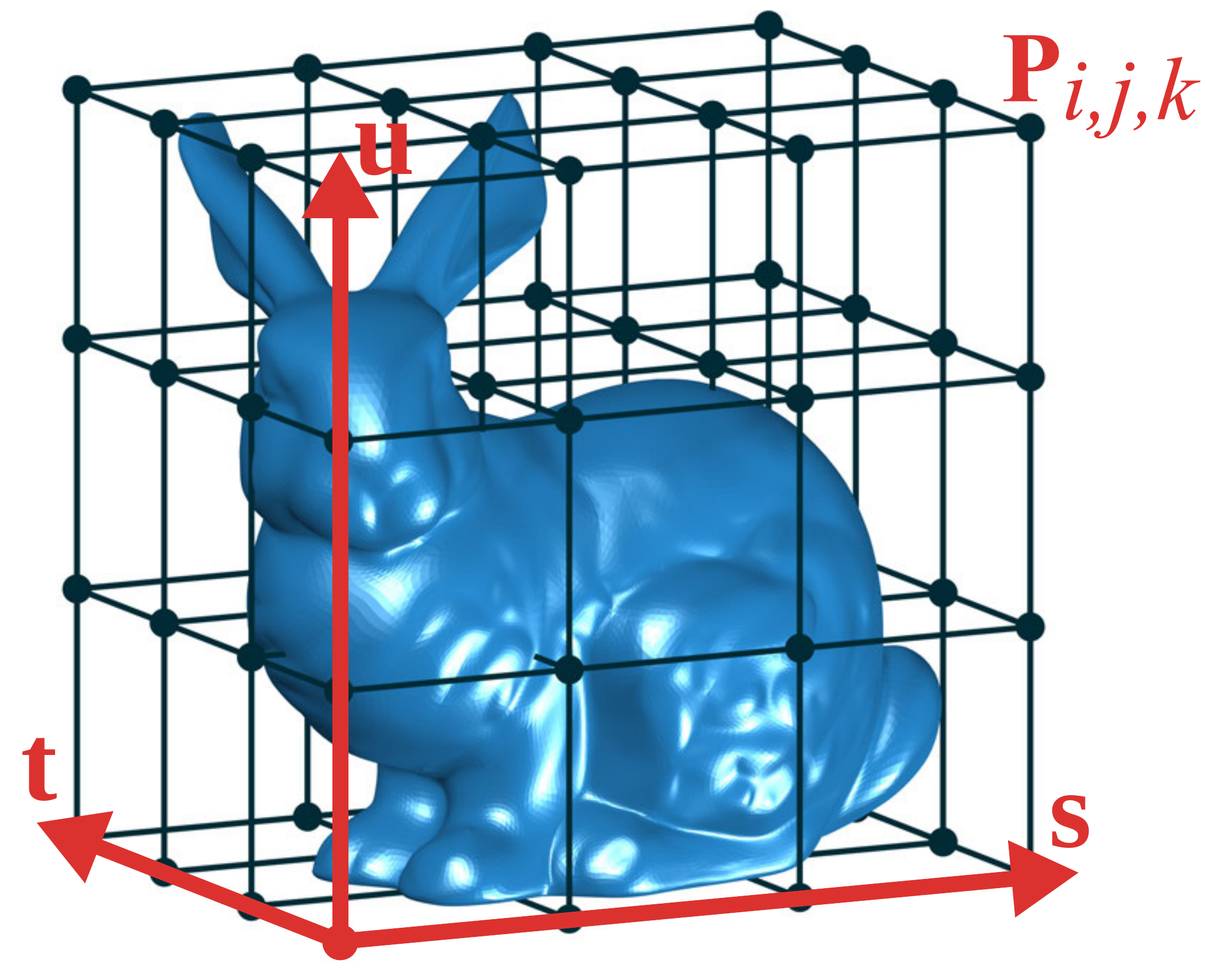}
      	\label{fig:FFD1}}
	\subfigure[]{
      	\includegraphics[width=100cm,height=3.6cm,keepaspectratio]{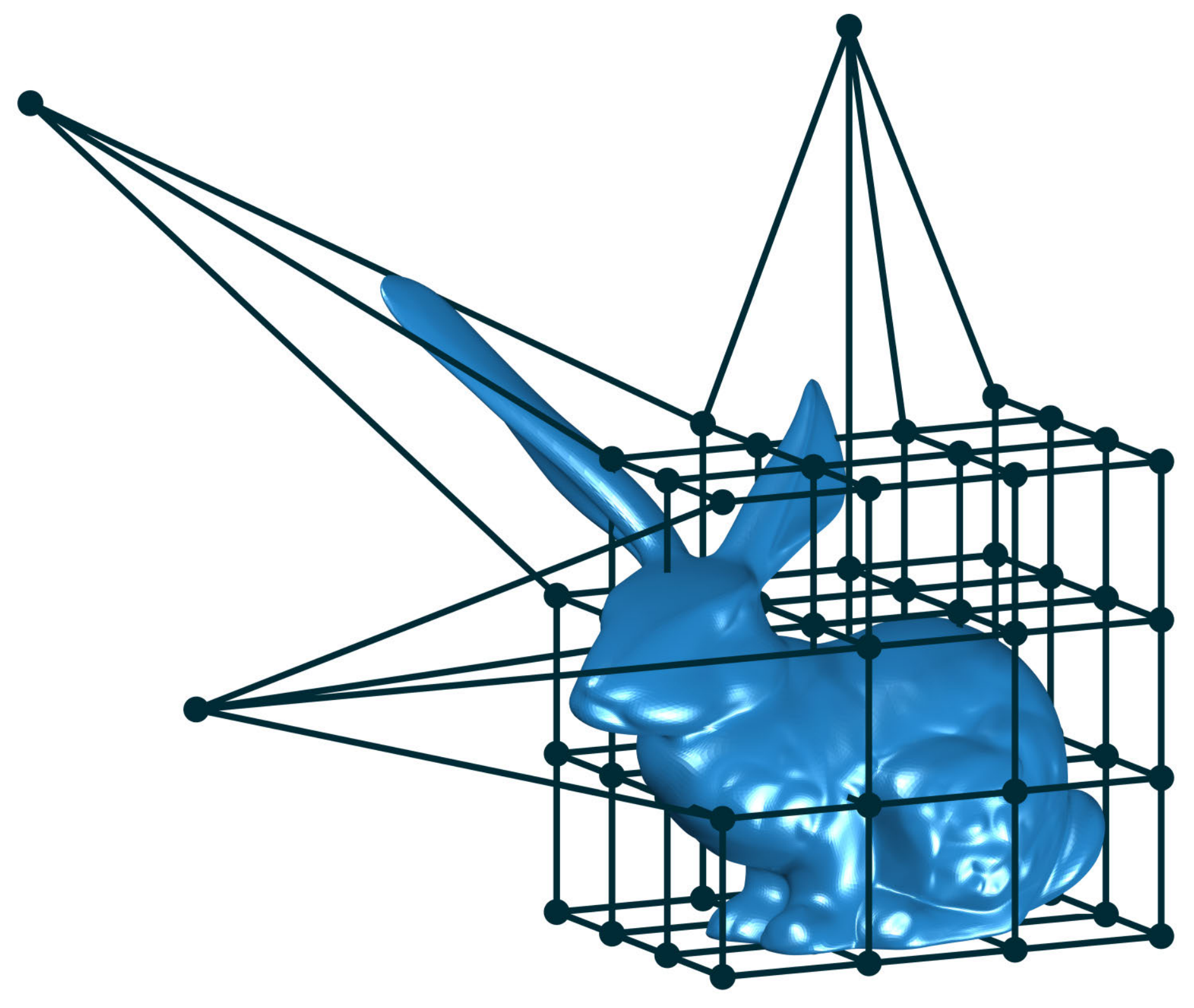}
    	\label{fig:FFD2}}
\caption{\textsc{Ffd} lattice of control points with $l,m,n {=} 3$: (a) The local coordinate system $(\mathbf{s},\mathbf{t},\mathbf{u})$ and the control points $\mathbf{P}_{i,j,k}$; (b) The Bunny's head deformation by free pulling three control points.}
\label{fig:FFD_bunny}
\end{figure}

\textsc{Ffd} can be thought as a virtual sculpting tool where the 3D mesh we wish to deform is first embedded into a lattice of control points. The object is then deformed by simply free manipulating these control points.

Formally, the \textsc{Ffd} method creates a parallelepiped-shaped lattice of control points with axes defined by the orthogonal vectors $\mathbf{s}, \mathbf{t}$ and $\mathbf{u}$ \cite{Sederberg1986}. The control points are defined by $l,m$ and $n$ which ``cut" the lattice in $l+1, m+1, n+1$ planes in the $\mathbf{s},\mathbf{t},\mathbf{u}$ directions respectively. Then, it is imposed a local coordinate for each object's vertex. An example is shown in Figure~\ref{fig:FFD1}.

In our implementation, we perform the deformation (see an example in Figure~\ref{fig:FFD2}) through a trivariate Bernstein tensor product as in \cite{Sederberg1986}, which is basically a weighted sum of the control points. It can also be formulated in terms of other blending functions such as B-splines. The deformed position, $\mathbf{x}_\mathit{ffd}$, of any arbitrary point, is given by, 

\startcompact{small}
\begin{equation}
	\mathbf{x}_\mathit{ffd}(s,t,u) = \sum^l_{i=0} \sum^m_{j=0} \sum^n_{k=0}B_{i,l}(s)B_{j,n}(t)B_{k,m}(u)\mathbf{P}_{i,j,k},
 	\label{eq:FFD3}
\end{equation}
\stopcompact{small}

\noindent where $\mathbf{x}_\mathit{ffd}$ contains the coordinates of the displaced point, $B_{\theta,n}(x)$ is the Bernstein polynomial of degree $n$ that sets the influence of each control point on every object's vertex, and $\mathbf{P}_{i,j,k}$ is the $i,j,k$-th control point. Equation~\eqref{eq:FFD3} is a linear function of $\mathbf{P}$ and it is written in a matrix form as,

\startcompact{small}
\begin{equation}
	\mathbf{X}_\mathit{ffd} = \mathbf{BP},
 	\label{eq:FFD6}
\end{equation}
\stopcompact{small}

\noindent where $\mathbf{X}_\mathit{ffd} \in \mathbb{R}^{N \times 3}$ are the vertices of the 3D mesh, $\mathbf{B} \in \mathbb{R}^{N \times M}$ is the deformation matrix, $\mathbf{P} \in \mathbb{R}^{M \times 3}$ are the control point coordinates, and $N$ and $M$ are the number of vertices and control points respectively.

\subsection{Dense Correspondences}
The majority of 3D models available do not have the same number of vertices or the same topology. Even so, this knowledge does not allow us to create dictionaries to deform a model by linear combination. To overcome this issue we find dense correspondences from a source model $\mathcal{S}$ to a target $\mathcal{T}$ using nonrigid ICP as in \cite{Chen2017} where for each vertex of $\mathcal{S}$ is found a corresponding point on the surface of $\mathcal{T}$. It uses a locally affine regularisation which assigns an affine transformation to each vertex and minimises the difference in the transformation of neighbouring vertices.

Consider we want to establish dense correspondences from $\mathcal{S}(\mathbf{\mathbf{V}_\mathcal{S}},\mathbf{\mathbf{E}_\mathcal{S}})$ to $\mathcal{T}(\mathbf{\mathbf{V}_\mathcal{T}},\mathbf{\mathbf{E}_\mathcal{T}})$ where $\mathbf{\mathbf{V}}, \mathbf{\mathbf{E}}$ are the vertices and edges (\ie connections) respectively, the source $\mathcal{S}$ is warped to the target $\mathcal{T}$ by nonrigid ICP such that the warped source model $\mathcal{S'}(\mathbf{\mathbf{V}'},\mathbf{\mathbf{E}_\mathcal{S}})$ can represent the target. The problem of nonrigid ICP is that if the source is not similar to the target, the warped model will be of low quality. This is the case when building up the graph with 3D models with great variations. To surpass this drawback we propose to first deform the source to fit the target using \textsc{Ffd} and then we apply nonrigid ICP to find the dense correspondences. The following convex optimisation problem is proposed to deform the source to the target in a free-form manner,

\startcompact{small}
\begin{equation}
	\begin{aligned}
         \argmin{\mathbf{\Delta p}}
		 \frac{1}{2} \| \mathbf{y} - (\mathbf{B}_\mathcal{A} \otimes \mathbf{I}_3)(\mathbf{p} + \mathbf{\Phi}\mathbf{\Delta p}) \|_2^2
         + \frac{\gamma}{2} \| \mathbf{\Phi}\mathbf{\Delta p} \|_2^2.
	\end{aligned}
\label{eq:icp_fit}
\end{equation}
\stopcompact{small}

\noindent The first term is a free-form 3D-3D registration where $\mathbf{y} \in \mathbb{R}^{3P}$ is the vectorized form\footnote{We use the vectorized form since we want to control the symmetry of the $x-, y-$ and $z-$direction of the \textsc{Ffd} grid $\mathbf{P}$ independently through the permutation matrix $\mathbf{\Phi}$.} of the target 3D anchors $vec([\mathbf{V}_\mathcal{T}]_\mathcal{A})$\footnote{$vec(\cdot)$ is the vectorization operator.}, $[\mathbf{V}_\mathcal{T}]_\mathcal{A} \in \mathbb{R}^{3 \times P}$, $P$ is the number of anchors, $\mathcal{A}$ is the set of 3D anchor indices, $\mathbf{B}_\mathcal{A}$ are the Bernstein basis of the set of 3D anchors, $\otimes$ is the Kronecker product, $\mathbf{p} \in \mathbb{R}^{3M}$ is the vectorized form of the control points $vec(\mathbf{P}^\top)$, $M$ is the number of control points, $\mathbf{\Delta p} \in \mathbb{R}^{3M}$ are the control point displacements to be optimized, and $\mathbf{\Phi} \in \mathbb{R}^{3M \times 3M}$ is a matrix to impose symmetry in the \textsc{Ffd} grid along the $x-$direction. In other words, only half of the control points are optimized while the other half is forced to be symmetric. If $\mathbf{\Phi}$ is the identity matrix, all control points can move freely. The second term is an $L^2$ regularization and $\gamma$ is the penalty weight. We penalize the $L^2$ norm of the control points displacement $\mathbf{\Delta p}$ to avoid bigger and unrealistic deformations that can be caused by the displacement of the grid far from the object. The problem can be efficiently solved by steepest descent. After solving this problem to make the source more similar to the target we can apply the nonrigid ICP to find the dense correspondences as in \cite{Chen2017}.
\subsection{Metrics to Create the Graph}
\label{metrics_graph}
To evaluate and define the similarity between the warped source model and the target we propose the use of two metrics. The first is a symmetric surface distance metric, $s_{dist}$, computed by densely sampling points on the faces and using normalized points distance to give us an estimate of the model similarity. It is defined as,

\startcompact{small}
\begin{equation}
	\begin{aligned}
		s_{dist} = \frac{1}{| \mathbf{V'} |} \sum_{\mathbf{v_i} \in \mathbf{V'}} f(\mathbf{v_i}, \mathcal{S_\mathcal{T}};\theta) + \frac{1}{| \mathbf{V_\mathcal{T}} |} \sum_{\mathbf{v_i} \in \mathbf{V_\mathcal{T}}} f(\mathbf{v_i}, \mathcal{S'};\theta), 
	\end{aligned}
\label{eq:surf_dist}
\end{equation}
\stopcompact{small}

\noindent where 

\startcompact{small}
\begin{equation}
	\begin{aligned}
	f(\mathbf{v}, \mathcal{S};\theta) = \begin{cases}
			1 & \quad \text{if } dist(\textbf{v}, \mathcal{S}) > \theta \\
			0 & \quad \text{otherwise.}
		\end{cases}
	\end{aligned}
\label{eq:function_f}
\end{equation}
\stopcompact{small}

\noindent The threshold $\theta$ is chosen through cross-validation such that consistent correspondences are found across object classes. This metric works fine if we consider the vertices without the edges. However, if we consider the edges the topology of the object may look incorrect (more details in the experiments section). To cope with that, we propose a second metric which is the intersection over union (IoU) between the voxel models defined as,

\startcompact{small}
\begin{equation}
	\begin{aligned}
		s_{IoU} = \frac{\mathcal{V'} \cap \mathcal{V_\mathcal{T}}}{\mathcal{V'} \cup \mathcal{V_\mathcal{T}}},
	\end{aligned}
\label{eq:iou}
\end{equation}
\stopcompact{small}

\noindent where $\mathcal{V'}$ and $\mathcal{V_\mathcal{T}}$ are the voxel models of the warped and target models respectively. We expect lower values for the surface distance score and higher values for the voxel IoU score. Once we find correspondences for every pair of 3D models from a specific class we define if they will be edges of the graph $\mathcal{G}$ according to both metrics, $s_{dist} < \theta_{dist}$ and $s_{IoU} > \theta_{IoU}$, where $\theta_{dist}$ and $\theta_{IoU}$ are predefined thresholds. The case where the warped model (\ie edge of $\mathcal{G}$) is poor we simply trim it off. Therefore, the graph is not fully connected but sparse allowing us to perform linear combinations to deform a 3D object in every subgraph. More specifically, denote $\Omega$ as an index set of the nodes in a certain subgraph with $\mathcal{S}_c(\mathbf{V}_c,\mathbf{E}_c)$ as the main node. There will always exist dense correspondences for all $i \in \Omega$ that allow us to deform a 3D model $\mathcal{S}(\mathbf{V},\mathbf{E})$ by linear combination,

\startcompact{small}
\begin{equation}
	\begin{aligned}
		\mathbf{V} = \alpha_c \mathbf{V}_c + \sum_{i \in \Omega} \alpha_i \mathbf{V}_c^i, \quad \mathbf{E} = \mathbf{E}_c,
	\end{aligned}
\label{eq:linearComb}
\end{equation}
\stopcompact{small}

\noindent where $\mathbf{V}$, $\mathbf{E}$ are matrices containing the vertices and edges respectively and $\alpha$'s are weights for the linear combination.

\section{Single Image 3D Reconstruction}
Once we build up the graph we can perform 3D reconstruction from a single image given the 2D anchors and the silhouette. The first step is the selection of a 3D model from the graph that best matches the image, and then we can refine the model using its correspondences which allows us to deform it by linear combinations. 

\subsection{Selecting a 3D Model}
If we are given a set of several 3D models and we are asked to find the best 3D model that fits a natural image we are likely to perform poorly. Humans can be good at annotating image characteristics, for example, anchors and silhouettes. However, when we have to look at a 3D model an judge if it fits a 2D image by estimating its projection within the scene, it turns out to be a hard task, especially when we have many possible options for the 3D model. In this work we propose to automatically select a 3D model from the graph previously described that best fits a 2D image given its anchors and silhouette. 

Given the graph $\mathcal{G}$ and a single image $\mathbf{I}$ with corresponding anchors $\mathbf{W} \in \mathbb{R}^{2 \times P}$, where $P$ is the number of anchors, and silhouette $\mathbf{\mathcal{S}}$ as a binary mask, we propose an optimization problem to register the 3D model anchors to the 2D anchors by allowing \textsc{Ffd}. Some anchors may be occluded so we denote an index set $\mathbf{\mathcal{L}}$ to indicate the anchors' visibility. 
For simplicity we employ an orthographic projection. We denote $\mathbf{R} \in \mathbb{R}^{2 \times 3}$ as the first two rows of a rotation matrix, $\mathbf{t} \in \mathbb{R}^{2 \times 1}$ as the translation vector, and $s$ as the scale of the camera model. The key idea is to simultaneously refine the camera pose and register the 3D anchors by displacing the control points of the \textsc{Ffd} lattice. 
We formulate this problem in the following way,

\startcompact{small}
\begin{equation}
	\begin{aligned}
        & \argmin{\mathbf{\Delta p}, \mathbf{R}, s, \mathbf{t}} \frac{1}{2} \| \mathbf{w}_\mathcal{L} - \big( (\mathbf{B}_\mathcal{L} \otimes s\mathbf{R})(\mathbf{p} + \mathbf{\Phi}\mathbf{\Delta p})+ \mathbf{t} \big) \|_2^2 \\
        & + \frac{\gamma}{2} \| \mathbf{\Phi} \mathbf{\Delta p} \|_2^2, \quad \text{subject to} \,\, \mathbf{RR}^\top = \mathbf{I}_2.
	\end{aligned}
\label{eq:FFD_PnP}
\end{equation}
\stopcompact{small}

\noindent The first term is the reprojection error where $\mathbf{w}_\mathcal{L}$ are the visible 2D anchors in the vectorized form, $\mathbf{B}_\mathcal{L}$ are the Bernstein basis of the set of visible 2D anchors, $\mathbf{p}$ is the vectorized form of the control points, $\mathbf{\Delta p}$ are the control point displacements, and $\mathbf{\Phi}$ is a matrix to impose symmetry in the \textsc{Ffd} grid. The second term is an $L^2$ regularization and $\gamma$ is the penalty weight as in Equation~\eqref{eq:icp_fit}. The objective function can be efficiently solved by the Alternating Direction Method of Multipliers (ADMM) \cite{Boyd2011}. We first introduce an auxiliary variable $\mathbf{Z}$ and rewrite the objective \eqref{eq:FFD_PnP} as follows,

\startcompact{small}
\begin{equation}
	\begin{aligned}
        & \argmin{\mathbf{\Delta p}, \mathbf{M}, \mathbf{Z}, \mathbf{t}} \frac{1}{2} \| \mathbf{w}_\mathcal{L} - \big( (\mathbf{B}_\mathcal{L} \otimes \mathbf{Z})(\mathbf{p} + \mathbf{\Phi}\mathbf{\Delta p}) + \mathbf{t} \big) \|_2^2 \\
		& + \frac{\gamma}{2} \| \mathbf{\Phi} \mathbf{\Delta p} \|_2^2, \quad \text{subject to} \,\, \mathbf{MM}^\top = s^2\mathbf{I}_2, \,\, \mathbf{M} = \mathbf{Z},
	\end{aligned}
\label{eq:FFD_PnP_aux}
\end{equation}
\stopcompact{small}

\noindent where $\mathbf{M}$ is the scaled rotation $s\mathbf{R}$. The augmented Lagrangian of our proposed objective \eqref{eq:FFD_PnP_aux} is formed as,

\startcompact{small}
\begin{equation}
	\begin{aligned}
        & \mathcal{L}_\rho(\mathbf{M},\mathbf{Z},\mathbf{\Delta p}, \mathbf{t}, \mathbf{\Lambda}) = \frac{1}{2} \| \mathbf{w}_\mathcal{L} - \big( (\mathbf{B}_\mathcal{L} \otimes \mathbf{Z})(\mathbf{p} + \mathbf{\Phi}\mathbf{\Delta p}) + \mathbf{t} \big) \|_2^2 \\
        & + \frac{\gamma}{2} \| \mathbf{\Phi} \mathbf{\Delta p} \|_2^2 + \langle \mathbf{\Lambda},\mathbf{M} - \mathbf{Z} \rangle_F + \frac{\rho}{2} \| \mathbf{M} - \mathbf{Z}\|^2_F,
	\end{aligned}
\label{eq:FFD_PnP_augm}
\end{equation}
\stopcompact{small}

\noindent where $\mathbf{\Lambda}$ is the dual variable, $\rho$ is a parameter controlling the step size in the optimization, and $\langle .,. \rangle_F $ is the Frobenius product of two matrices. The ADMM alternates the following steps until convergence:

\startcompact{small}
\begin{align}
	& \mathbf{M}^{k} = \argmin{\mathbf{M}} \mathcal{L}_\rho(\mathbf{M},\mathbf{Z}^{k-1},\mathbf{\Delta p}^{k-1}, \mathbf{t}^{k-1}, \mathbf{\Lambda}^{k-1}), \label{eq:1}\\
    & \text{subject to} \,\, \mathbf{MM}^\top = s^2\mathbf{I}_2; \nonumber \\
	& \mathbf{Z}^{k} = \argmin{\mathbf{Z}} \mathcal{L}_\rho(\mathbf{M}^k,\mathbf{Z},\mathbf{\Delta p}^{k-1}, \mathbf{t}^{k-1}, \mathbf{\Lambda}^{k-1}); \label{eq:2}\\
	& \mathbf{\Delta p}^{k} = \argmin{\mathbf{\Delta p}} \mathcal{L}_\rho(\mathbf{M}^k,\mathbf{Z}^k,\mathbf{\Delta p}, \mathbf{t}^{k-1}, \mathbf{\Lambda}^{k-1}); \label{eq:3}\\
	& \mathbf{t}^{k} = \argmin{\mathbf{t}} \mathcal{L}_\rho(\mathbf{M}^k,\mathbf{Z}^k,\mathbf{\Delta p}^k, \mathbf{t}, \mathbf{\Lambda}^{k-1}); \label{eq:4}\\
	& \mathbf{\Lambda}^{k} = \mathbf{\Lambda}^{k-1}  + \rho(\mathbf{M}^k - \mathbf{Z}^k) \label{eq:5}.
\end{align}
\stopcompact{small}

The subproblem \eqref{eq:1} is solved by,

\startcompact{small}
\begin{equation}
		\begin{aligned}
		& \argmin{\mathbf{M}} \mathcal{L}_\rho(\mathbf{M},\mathbf{Z}^{k-1},\mathbf{\Delta p}^{k-1}, \mathbf{t}^{k-1}, \mathbf{\Lambda}^{k-1}) = \\
       	& \mathbf{U}  
		\begin{bmatrix}
    		(\sigma_1 + \sigma_2)/2 &  \\
    		 & (\sigma_1 + \sigma_2)/2 
		\end{bmatrix}
        \mathbf{V}^\top,
	\end{aligned}
\label{eq:subproblem1}
\end{equation}
\stopcompact{small}

\noindent where $\mathbf{Z} - \frac{\Lambda}{\rho}  = \mathbf{U} \left[ \begin{smallmatrix} \sigma_1& \\  &\sigma_2 \end{smallmatrix} \right] \mathbf{V}^\top$, $\mathbf{U}$, $\mathbf{V}$ and $\sigma$ denote the left singular vectors, right singular vectors and the singular values of $\mathbf{Z} - \frac{\Lambda}{\rho}$ respectively.


The subproblem \eqref{eq:2} is solved by,

\startcompact{small}
\begin{equation}
	\begin{aligned}
		& \argmin{\mathbf{Z}} \mathcal{L}_\rho(\mathbf{M}^k,\mathbf{Z},\mathbf{\Delta p}^{k-1}, \mathbf{t}^{k-1}, \mathbf{\Lambda}^{k-1}) = \\
		& \big( (\mathbf{W}_\mathcal{L} - \mathbf{t})\mathbf{S}^\top + \mathbf{\Lambda} + \rho\mathbf{M} \big) (\mathbf{S}\mathbf{S}^\top + \rho \mathbf{I})^+,
	\end{aligned}
\label{eq:subproblem2}
\end{equation}
\stopcompact{small}

\noindent where $\mathbf{S} = unvec(\mathbf{p} + \mathbf{\Phi} \mathbf{\Delta p})\mathbf{B}_\mathcal{L}$\footnote{$unvec(\cdot)$ reshapes the argument to a matrix form.} $\in \mathbb{R}^{3 \times |\mathcal{L}|}$ and $|\mathcal{L}|$ is the number of visible anchors.

The subproblem \eqref{eq:3} is solved by,

\startcompact{small}
\begin{equation}
	\begin{aligned}
		& \argmin{\mathbf{\Delta p}} \mathcal{L}_\rho(\mathbf{M}^k,\mathbf{Z}^k,\mathbf{\Delta p}, \mathbf{t}^{k-1}, \mathbf{\Lambda}^{k-1}) = \\		
        & \big( \mathbf{\Phi}^\top(\mathbf{B}_\mathcal{L} \otimes \mathbf{Z})^\top(\mathbf{B}_\mathcal{L} \otimes \mathbf{Z})\mathbf{\Phi} + \gamma\mathbf{\Phi} \big)^+ \\ 
        & \big( \mathbf{\Phi}^\top(\mathbf{B}_\mathcal{L} \otimes \mathbf{Z})^\top  \big( \mathbf{w}_\mathcal{L} - (\mathbf{B}_\mathcal{L} \otimes \mathbf{Z})\mathbf{p} + \mathbf{t} \big) \big).
	\end{aligned}
\label{eq:subproblem3}
\end{equation}
\stopcompact{small}

The subproblem \eqref{eq:4} is solved by,

\startcompact{small}
\begin{equation}
	\begin{aligned}
		& \argmin{\mathbf{t}} \mathcal{L}_\rho(\mathbf{M}^k,\mathbf{Z}^k,\mathbf{\Delta p}^{k}, \mathbf{t}, \mathbf{\Lambda}^{k-1}) =  \frac{\sum_{l \in \mathcal{L}} ( \mathbf{W}_l - \mathbf{ZS}_l )}{|\mathcal{L}|}.
	\end{aligned}
\label{eq:subproblem4}
\end{equation}
\stopcompact{small}

This problem optimizes the projection of the 3D anchors to the 2D anchors by allowing \textsc{Ffd}. It is analogous to the Perspective-n-Point (PnP) algorithm but allowing free-form deformation for a better 3D-2D fitting. We exhaustively search over all 3D models in the graph and then select the one with the highest IoU. The IoU is computed between the input image silhouette and the silhouette generated from the output deformed 3D model.

\begin{figure*}[ht!]
	\centering
    \subfigure[Source]{
       	\includegraphics[trim={11cm 5cm 11cm 4cm},clip,height=2.2cm,keepaspectratio]{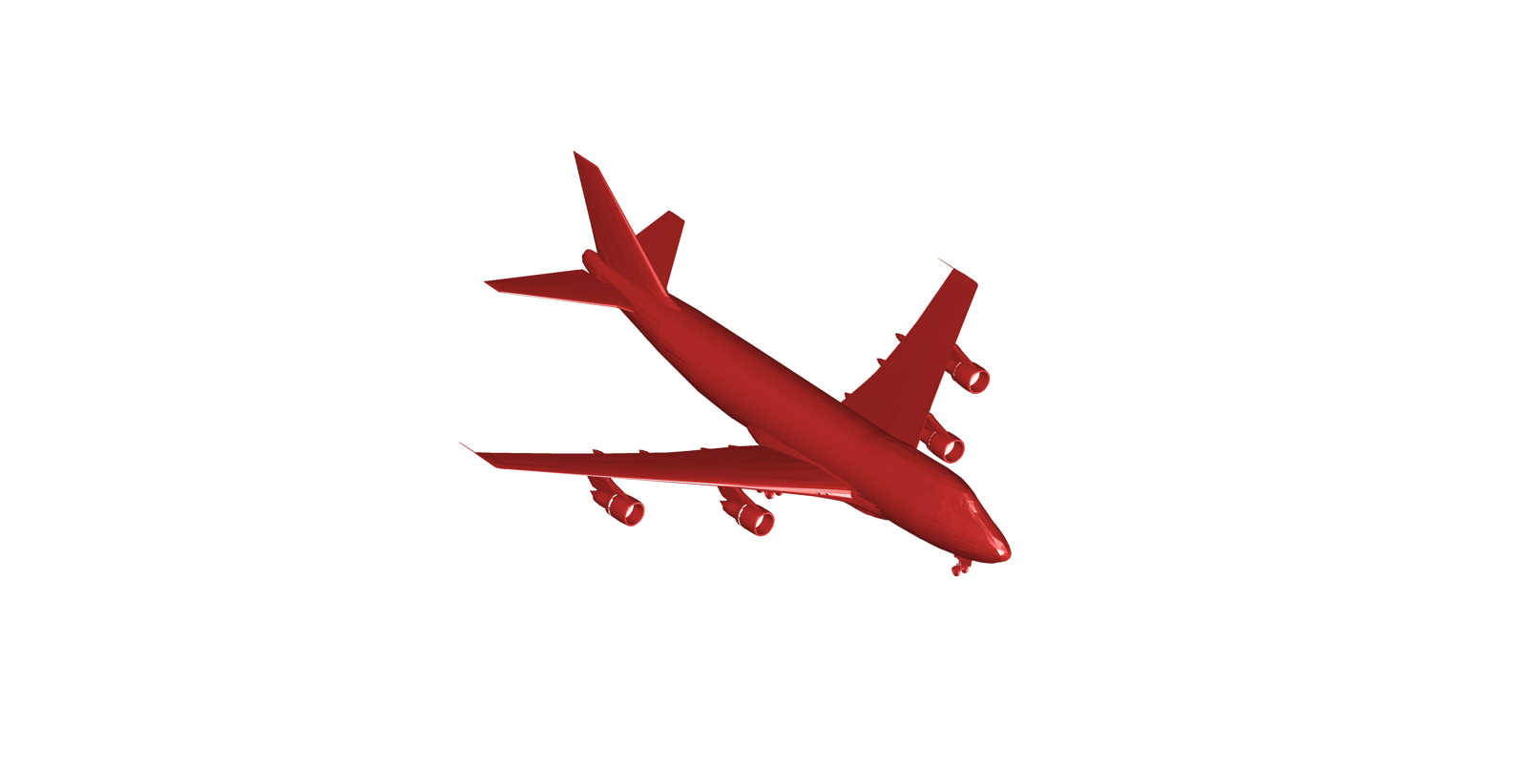}
      	\label{fig:nricp_plane_1}}
	\subfigure[Target]{
      	\includegraphics[trim={11cm 5cm 9cm 4cm},clip,height=2.2cm,keepaspectratio]{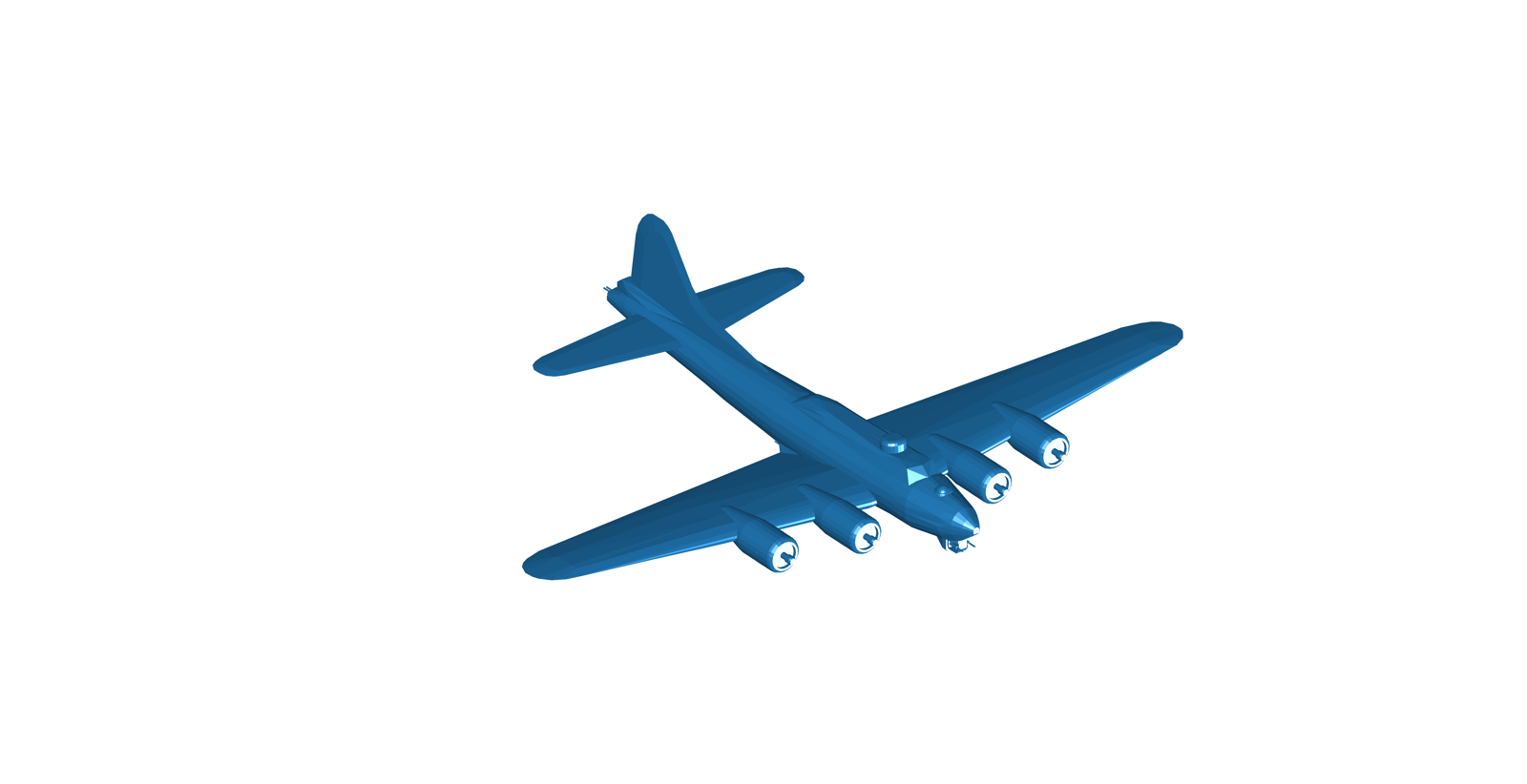}
    	\label{fig:nricp_plane_2}}
   	\subfigure[Nonrigid ICP]{
      	\includegraphics[trim={11cm 5cm 9cm 5cm},clip,height=2.2cm,keepaspectratio]{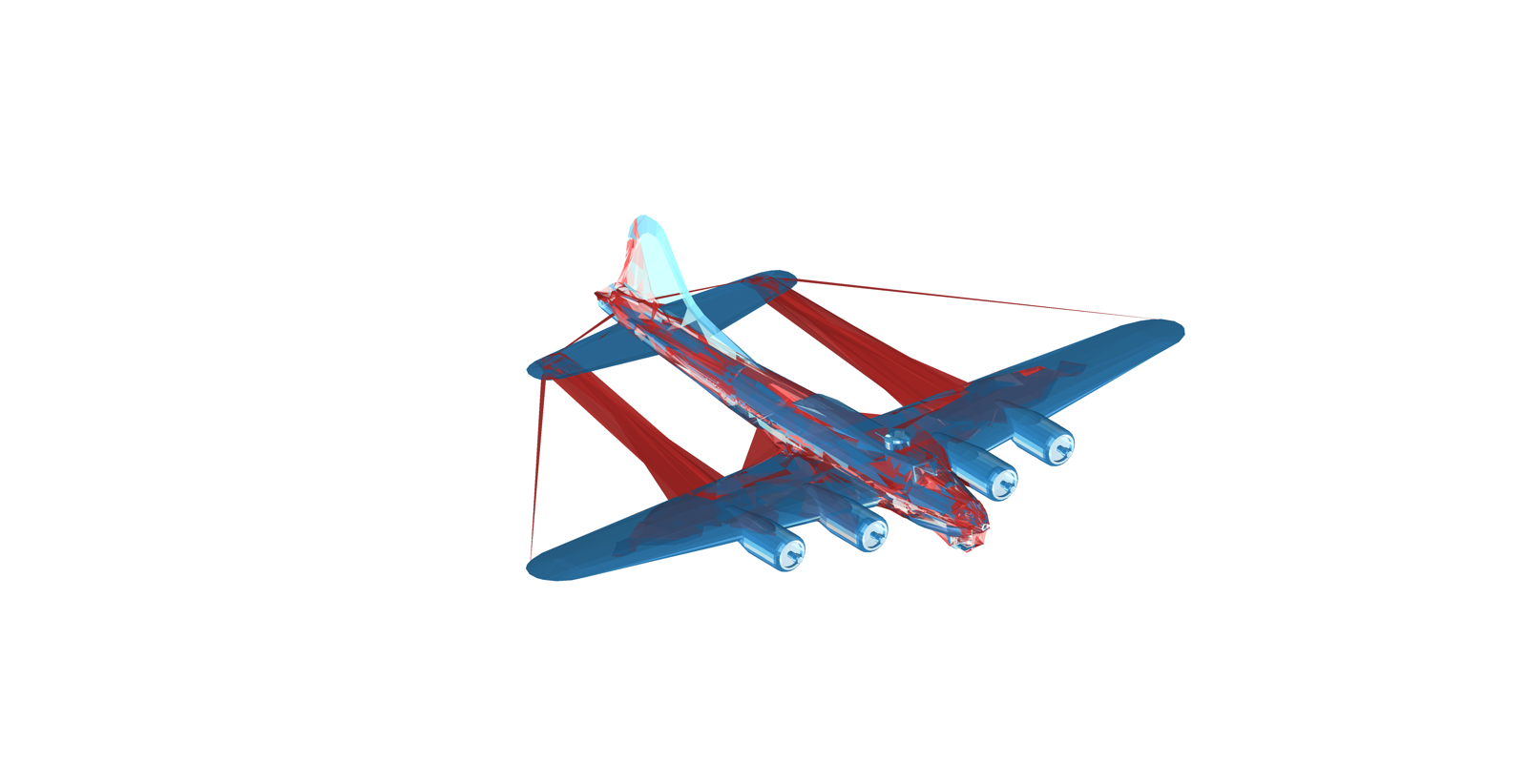}
    	\label{fig:nricp_plane_3}}
   	\subfigure[Warped model \cite{Chen2017}]{
      	\includegraphics[trim={11cm 5cm 9cm 5cm},clip,height=2.2cm,keepaspectratio]{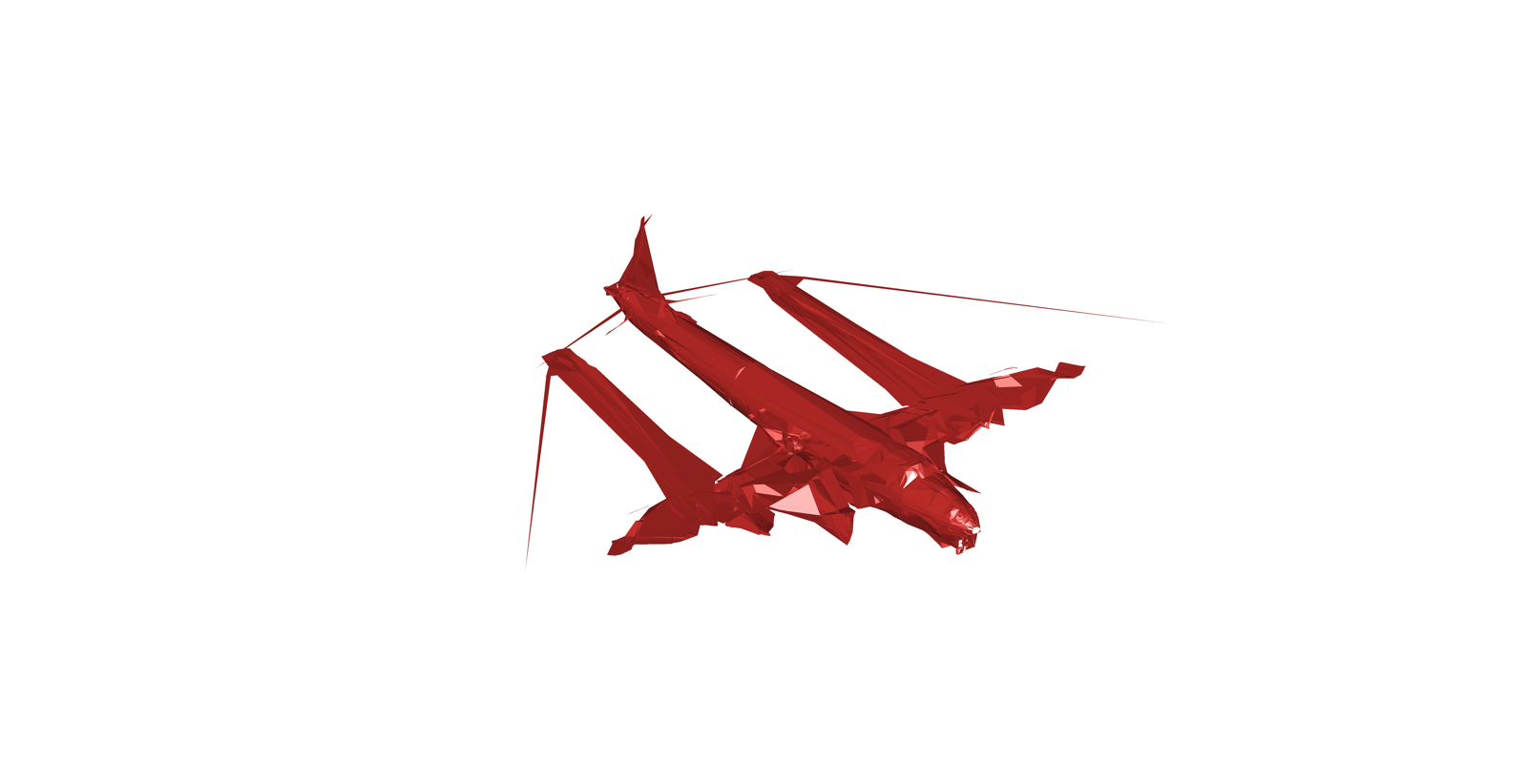}
    	\label{fig:nricp_plane_4}}
   	\subfigure[FFD on source]{
      	\includegraphics[trim={3cm 2.5cm 2cm 2cm},clip,height=2.5cm,keepaspectratio]{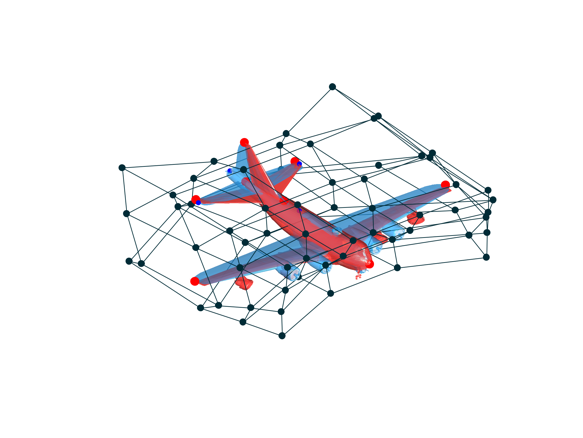}
    	\label{fig:nricp_plane_5}}
   	\subfigure[Deformed source]{
      	\includegraphics[trim={11cm 5cm 9cm 4cm},clip,height=2.2cm,keepaspectratio]{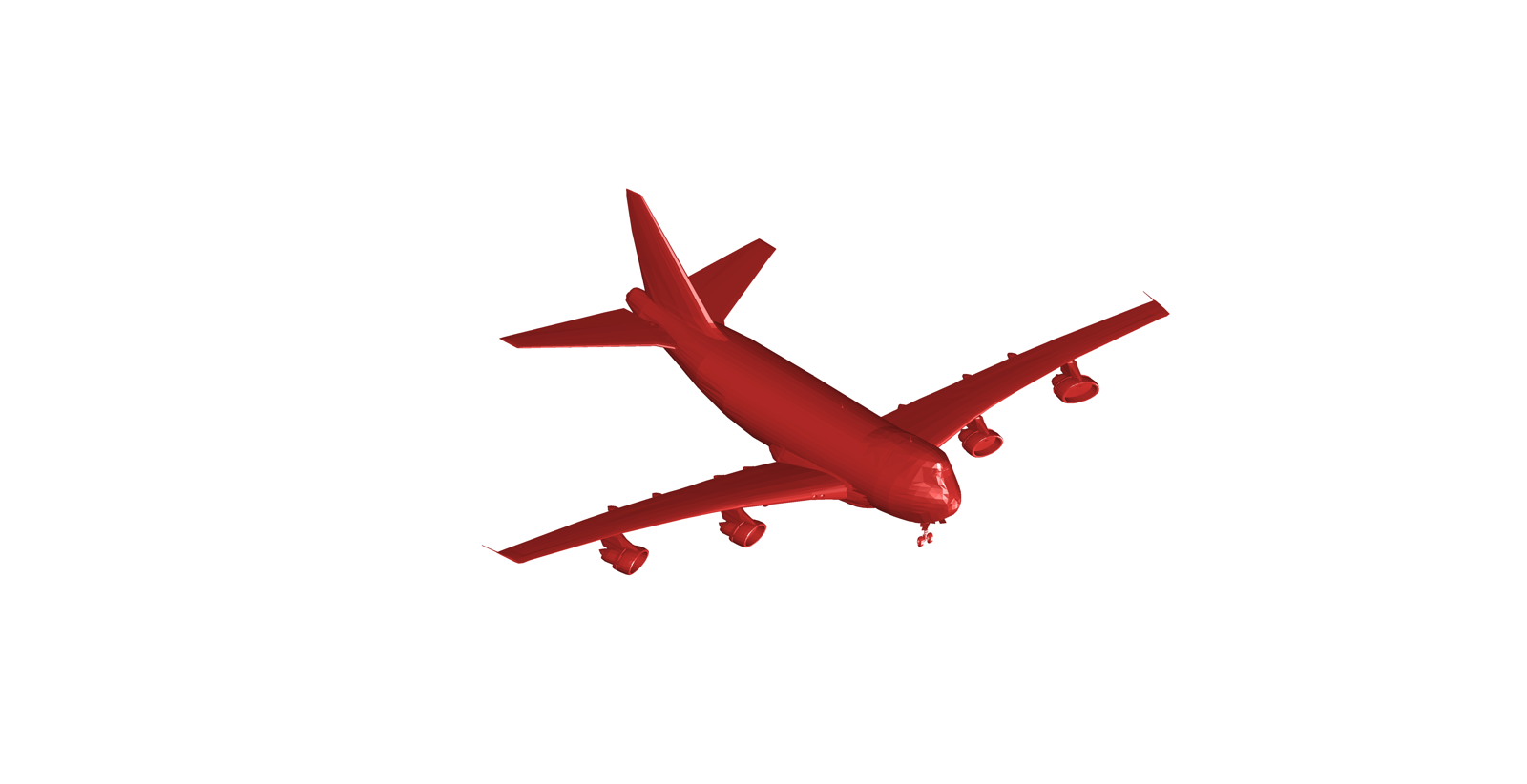}
    	\label{fig:nricp_plane_6}}
   	\subfigure[Nonrigid ICP]{
      	\includegraphics[trim={11cm 5cm 9cm 4cm},clip,height=2.2cm,keepaspectratio]{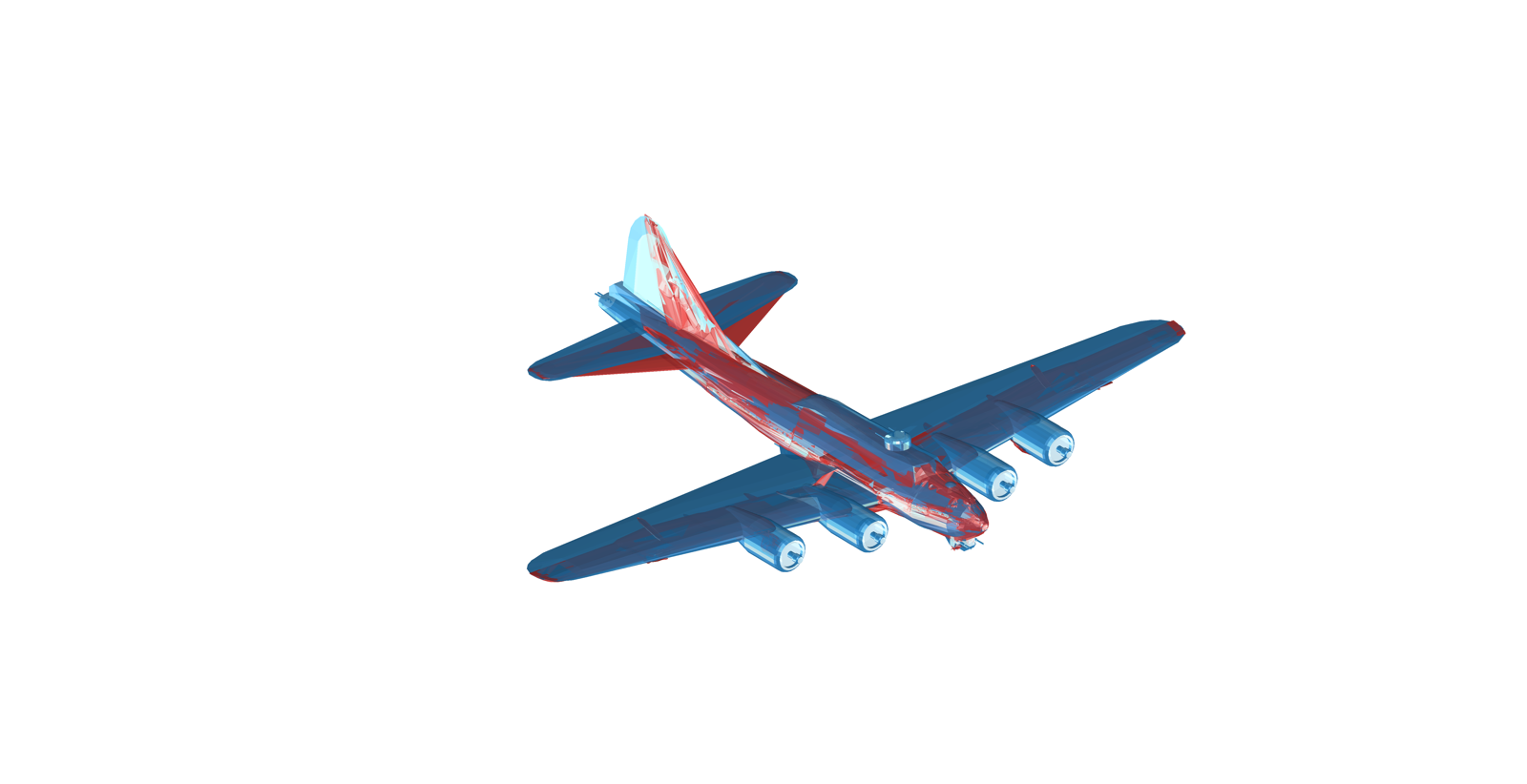}
    	\label{fig:nricp_plane_7}}
   	\subfigure[Our warped model]{
      	\includegraphics[trim={11cm 5cm 9cm 4cm},clip,height=2.2cm,keepaspectratio]{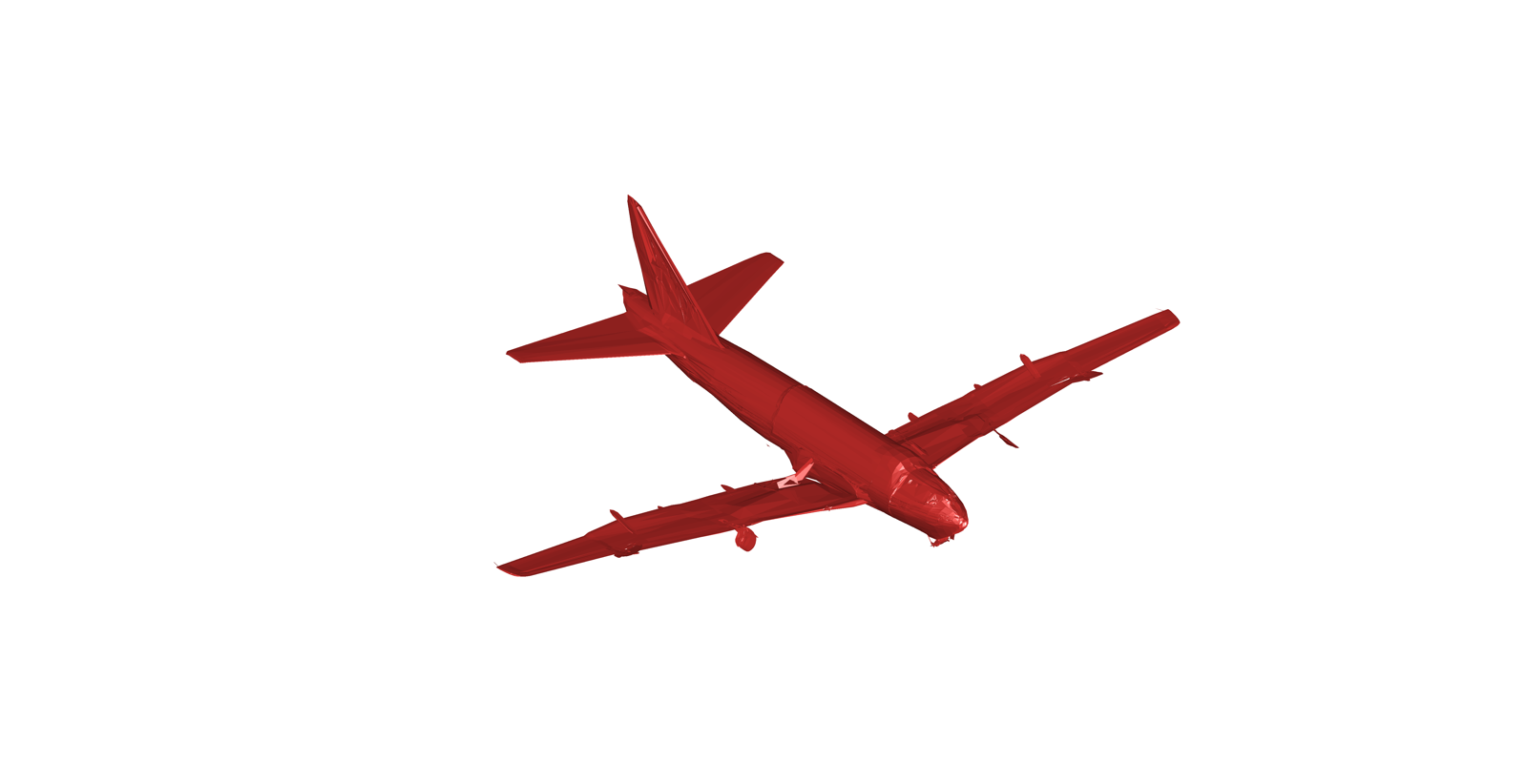}
    	\label{fig:nricp_plane_8}}
\caption{Finding dense correspondences. In the first row the source aeroplane (red) is warped to fit the target (blue) by using nonrigid ICP as proposed in \cite{Chen2017}. In the second row it shows our method that first brings the two aeroplanes closer to each other by deforming the source using \textsc{Ffd} to fit the target. Then, we apply nonrigid ICP between the deformed source and the target to find dense correspondences.}
\label{fig:nricp_plane}
\end{figure*}

\subsection{Refining the 3D Model}
Once a 3D model candidate is selected (\ie a node from $\mathcal{G}$) we can refine both the camera pose and the 3D model by using its edges to perform linear combination. The idea is to simultaneously refine the camera pose and the 3D anchors while forcing the vertices of the 3D model to be projected inside the image silhouette $\mathcal{S}$ by optimizing its combination weights $\alpha$'s. Let $\mathcal{S}_c(\mathbf{V}_c,\mathbf{E}_c)$ denote the selected subgraph (\ie the node/3D model) and $\Omega$ an index set of the nodes in the subgraph. The deformable model $\mathcal{S}(\mathbf{V},\mathbf{E})$ is represented by $\mathbf{X} = \alpha_c \mathbf{X}_c + \sum_{i \in \Omega} \alpha_i \mathbf{X}_c^i$, where $\mathbf{X}_c$, $\mathbf{X}_c^i$ are the 3D anchors of the selected 3D model $\mathcal{S}_c$ and $\mathcal{S}_c^i$ respectively, $i$ indicates the $i$-th node of the subgraph (\ie the dense correspondences), and $\alpha$'s are the weights for the linear combination to be optimized. We propose the following optimization problem as in \cite{Chen2017},

\startcompact{small}
\begin{equation}
	\begin{aligned}
        & \argmin{s, \mathbf{R}, \alpha} \frac{1}{2} \sum_{l \in \mathcal{L}} \| \mathbf{w}_l - \big( s\mathbf{R}(\alpha_c [\mathbf{X}_c]_l + \sum_{i \in \Omega}\alpha_i[\mathbf{X}^i_c]_l) + \mathbf{t} \big) \|_2^2 \\
        & + \mu \sum_{l=1}^N\mathcal{C} \big( s\mathbf{R}(\alpha_c [\mathbf{V}_c]_l + \sum_{i \in \Omega}\alpha_i[\mathbf{V}^i_c]_l) + \mathbf{t} \big) + \frac{\gamma}{2} \sum_{i \in \Omega} \alpha_i^2,
	\end{aligned}
\label{eq:silh_fit}
\end{equation}
\stopcompact{small}

\noindent where $\mathbf{w}_l$ is the 2D position of the $l$-th visible anchor, for $l = 1,\ldots,P$, $N$ is the number of vertices, $\mathbf{\mathcal{C}}$ is the Chamfer distance map \cite{Barrow1977} of the image silhouette, and $\mu$, $\gamma$ are penalty weights. The first term is the reprojection error of the 3D anchors, the second term penalizes the vertices projected outside the silhouette to ensure silhouette consistency. The penalty is proportional to the Chamfer distance to the silhouette. The third term is an $L^2$ regularization to avoid larger deformations. The objective function can be efficiently solved by steepest descent. The gradient $\nabla_{\alpha_i}$ of the objective function with respect to $\alpha_i$ is,

\startcompact{small}
\begin{equation}
	\begin{aligned}
        & \nabla_{\alpha_i} = \sum_{l \in \mathcal{L}} \Big( \mathbf{w}_l - \big( s\mathbf{R}(\alpha_c [\mathbf{X}_c]_l + \sum_{i \in \Omega}\alpha_i[\mathbf{X}^i_c]_l) + \mathbf{t} \big) \Big)^\top s\mathbf{R} [\mathbf{X}_c^i]_l \\
        & + \mu \sum_{l=1}^N \nabla \mathcal{C}^\top s\mathbf{R}[\mathbf{V}_c^i]_l + \gamma\alpha_i,
	\end{aligned}
\label{eq:silh_fit3}
\end{equation}
\stopcompact{small}

\noindent where $\nabla \mathcal{C}$ is the derivative of the Chamfer distance. To optimize the rotation $\mathbf{R}$ we use the exponential map to transform the angle-axis rotation representation to a corresponding rotation matrix. Let $\mathbf{R}$ be equal to $\mathbf{R}e^{[\mathbf{\xi}]_\times}$ where $[\cdot]_\times$ is the skew-symmetric matrix. The gradient $\nabla_{\mathbf{\xi}}$ of the objective function with respect to $\mathbf{\xi}$ is,

\startcompact{small}
\begin{equation}
	\begin{aligned}
        & \nabla_{\mathbf{\xi}} = \sum_{l \in \mathcal{L}} \Big( \mathbf{w}_l - \big( s\mathbf{R}(\alpha_c [\mathbf{X}_c]_l + \sum_{i \in \Omega}\alpha_i[\mathbf{X}^i_c]_l) + \mathbf{t} \big) \Big)^\top \\
        &  \Big( s\mathbf{R}\frac{\partial[\mathbf{\xi}]_\times}{\xi_j}(\alpha_c[\mathbf{X}_c]_l + \sum_{i \in \Omega} \alpha_i[\mathbf{X}_c^i]_l) \Big) \\
        &  + \mu \sum_{l=1}^N \nabla \mathcal{C}^\top \Big( s\mathbf{R}\frac{\partial[\mathbf{\xi}]_\times}{\xi_j}(\alpha_c[\mathbf{V}_c]_l + \sum_{i \in \Omega} \alpha_i[\mathbf{V}_c^i]_l) \Big).
	\end{aligned}
\label{eq:silh_fit4}
\end{equation}
\stopcompact{small}

\noindent The gradient $\nabla_{\mathbf{t}}$ with respect to the translation $\mathbf{t}$ is,

\startcompact{small}
\begin{equation}
	\begin{aligned}
       \nabla_{\mathbf{t}} = \sum_{l \in \mathcal{L}} \Big( \mathbf{w}_l - \big( s\mathbf{R}(\alpha_c [\mathbf{X}_c]_l + \sum_{i \in \Omega}\alpha_i[\mathbf{X}^i_c]_p) + \mathbf{t} \big) \Big) + \mu \sum_{l=1}^N \nabla \mathcal{C}.
	\end{aligned}
\label{eq:silh_fit5}
\end{equation}
\stopcompact{small}

\noindent The scale $s$ is absorbed by the weights $\alpha$'s.

\section{Experiments}
Experiments were performed to evaluate how generalizable the graph structure is when compactly representing new 3D models and how expressive the 3D reconstructed models from single images are when using our method.

\textbf{Datasets.} For building up and evaluating the graph structure we used 3D models of eight classes from ShapeNet \cite{ShapeNet}. We used 3D anchors that were manually annotated by \cite{Chen2017}. For the evaluation of the 3D reconstruction from a single image we used the PASCAL3D+ dataset \cite{Pascal3D} since we have access to the camera pose and 3D model ground truths, 2D anchors, and silhouettes for several natural images in the eight classes chosen.

\textbf{Metrics.} To quantify the quality of the graph structure we used the metrics described in Subsection~\ref{metrics_graph} and for the 3D reconstruction task we used three metrics: \textit{(i)} 2D anchor reprojection error, $e_{RP}$, \textit{(ii)} camera pose error, $e_{pose}$, and \textit{(iii)} 3D structure error, $e_{3D}$. The 2D anchor reprojection error measures the accuracy of the reprojected anchors by computing the mean Euclidean distance between the projected anchors of the estimated model and the ground truth anchors. The camera pose error measures the accuracy of the estimated pose by computing the Frobenius norm of the difference between the estimated and the ground truth camera pose. The 3D structure error measures the quality of the reconstructed 3D model by comparing to its ground truth using the surface distance metric shown in Equation~\eqref{eq:surf_dist}.

\subsection{Creating the Graph}
\begin{figure*}[ht!]
	\centering
    \subfigure[$\mathcal{S}$]{
       	\includegraphics[trim={15cm 3.7cm 15cm 5cm},clip,height=2.2cm,keepaspectratio]{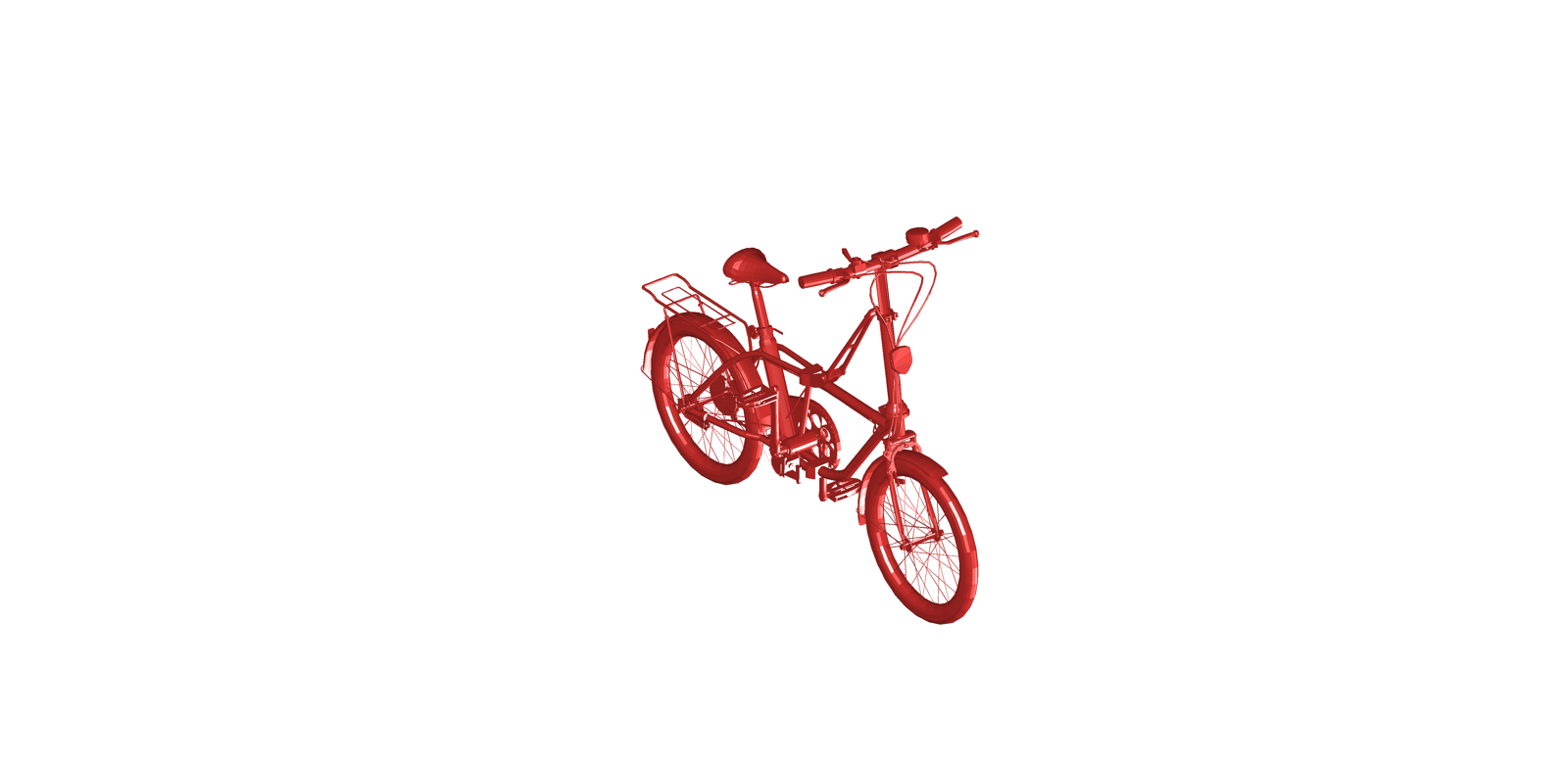}
      	\label{fig:bike_source}}
	\subfigure[$\mathcal{T}$]{
      	\includegraphics[trim={15cm 3.7cm 14cm 3.5cm},clip,height=2.2cm,keepaspectratio]{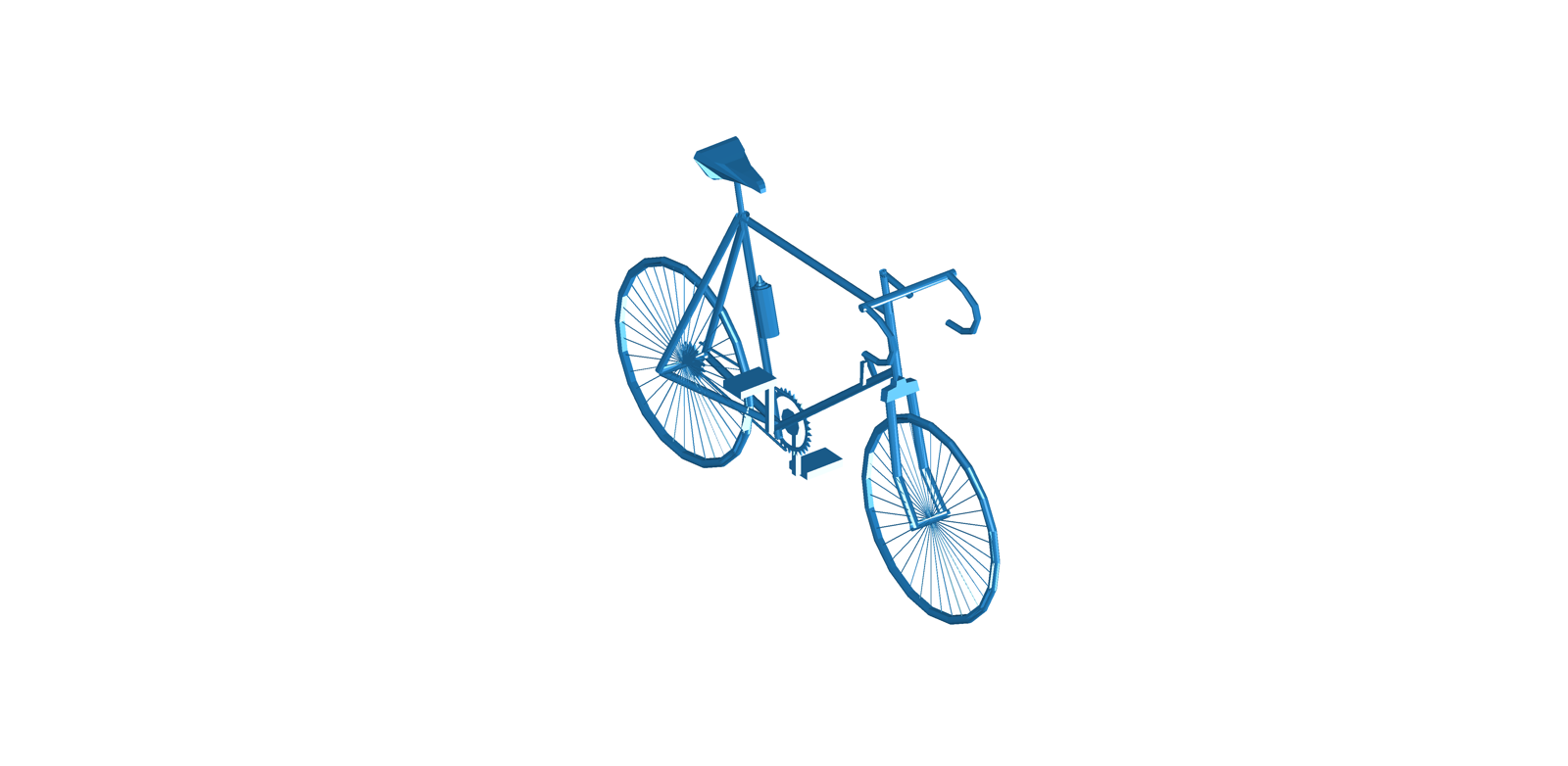}
    	\label{fig:bike_target}}
   	\subfigure[$\mathcal{T}$ point cloud]{
      	\includegraphics[trim={15cm 3.7cm 14cm 3.5cm},clip,height=2.2cm,keepaspectratio]{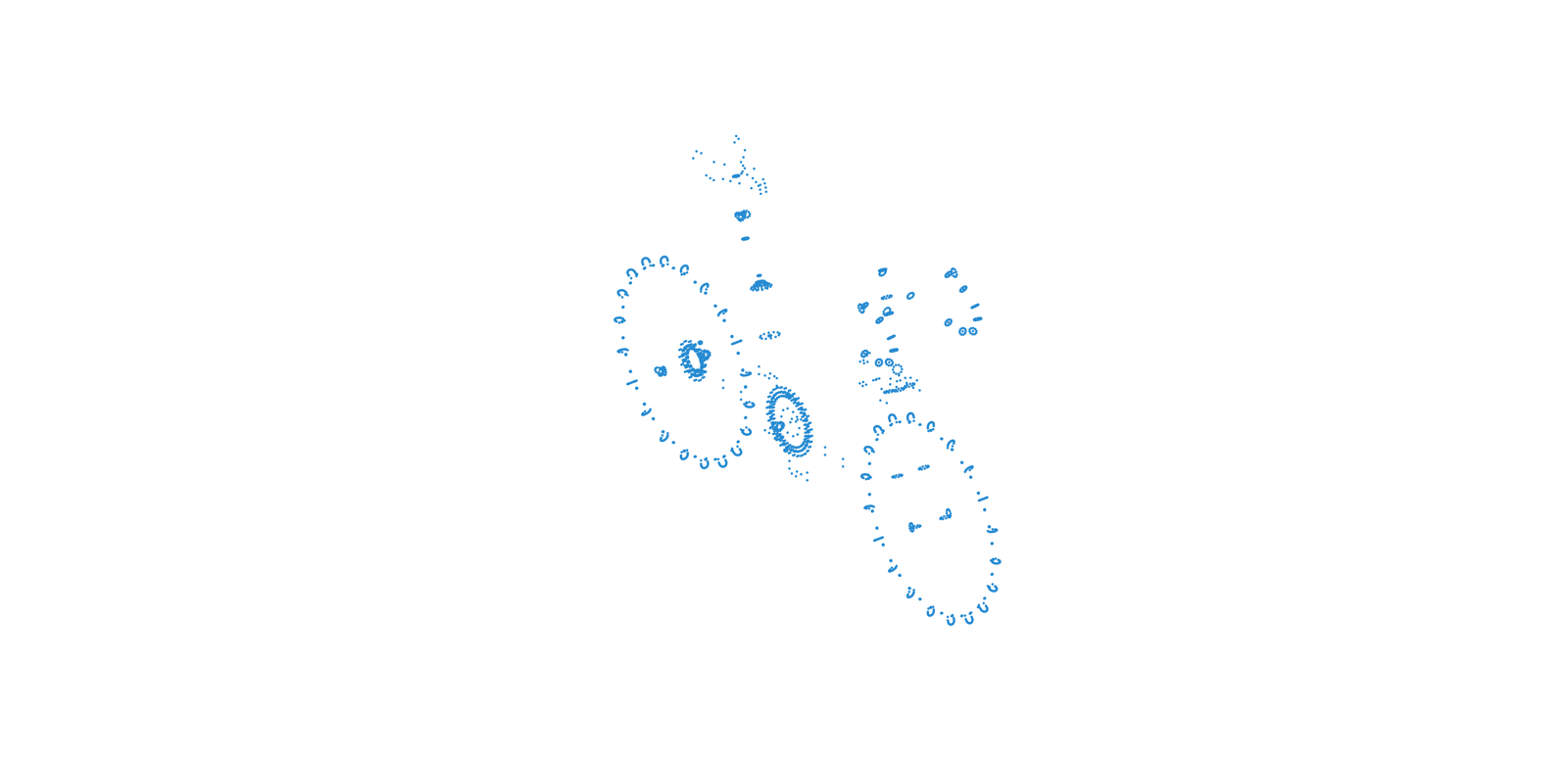}
    	\label{fig:bike_target_PC}}
   	\subfigure[$\mathcal{V}_\mathcal{T}$]{
      	\includegraphics[trim={18cm 7.5cm 16cm 7cm},clip,height=2.2cm,keepaspectratio]{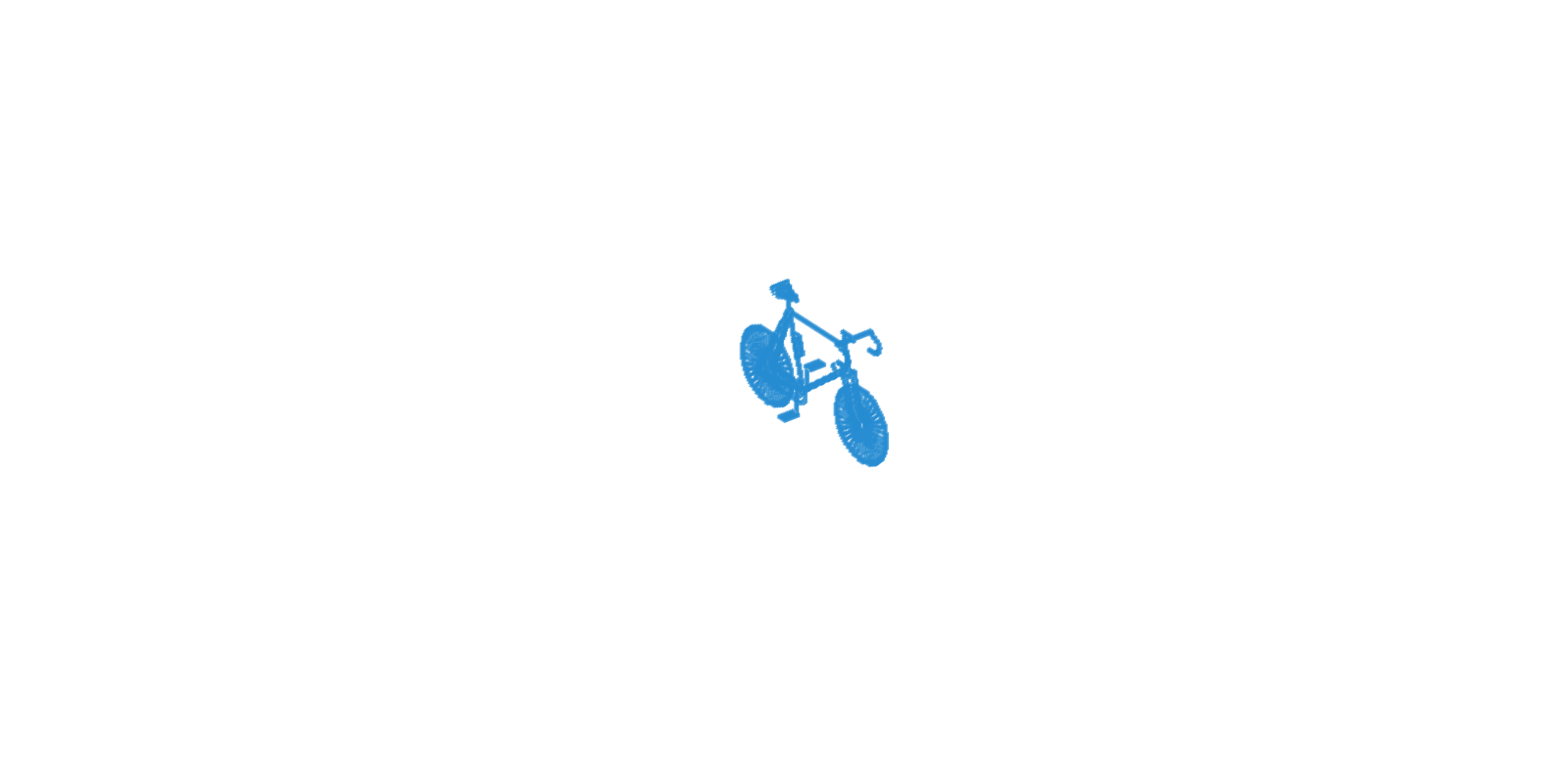}
    	\label{fig:bike_target_vxl}}
   	\subfigure[$\mathcal{S'}$]{
      	\includegraphics[trim={16cm 3.7cm 14cm 3.5cm},clip,height=2.2cm,keepaspectratio]{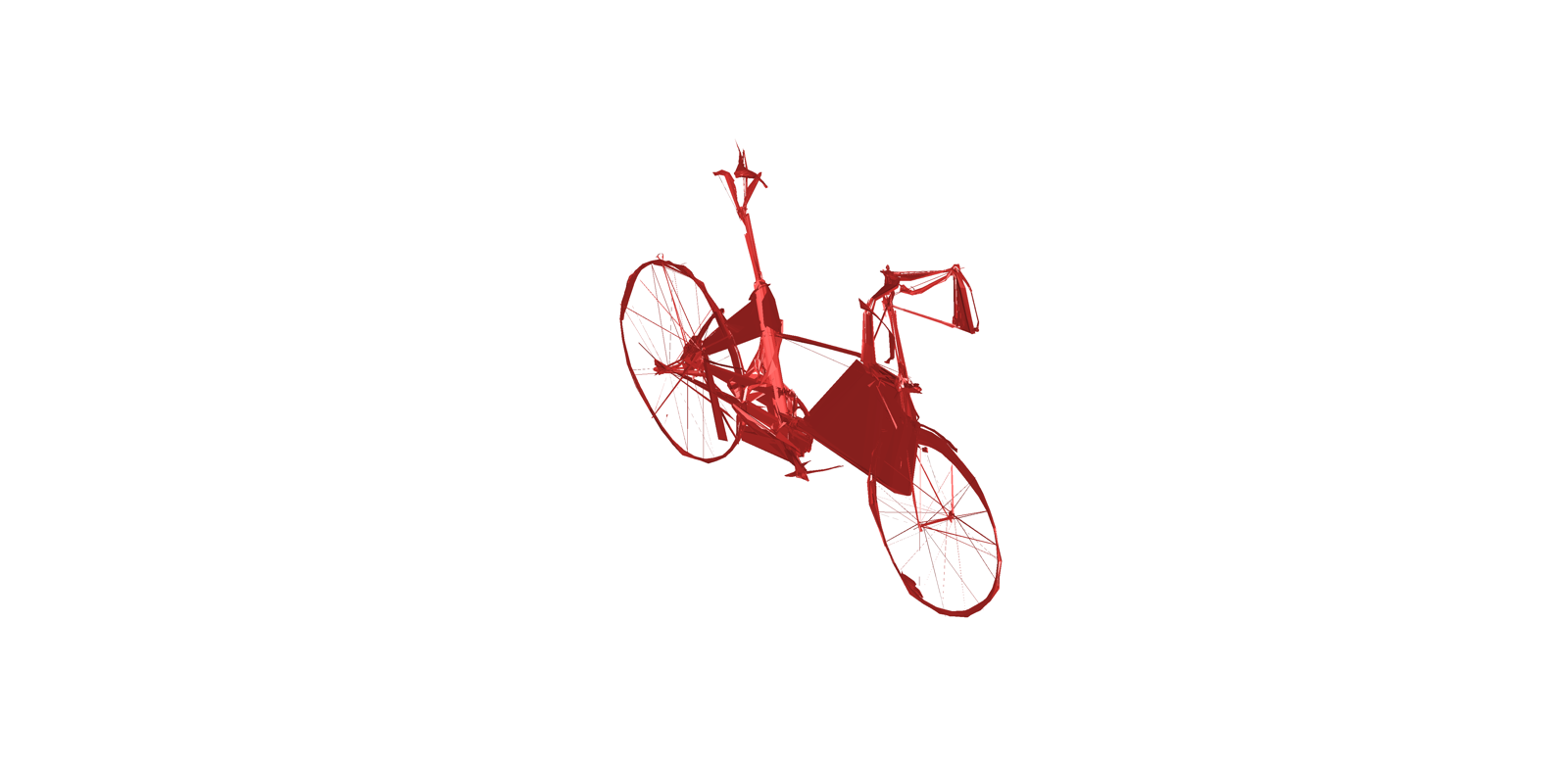}
    	\label{fig:bike_def}}
   	\subfigure[$\mathcal{S'}$ point cloud]{
      	\includegraphics[trim={15cm 3.7cm 14cm 3.5cm},clip,height=2.2cm,keepaspectratio]{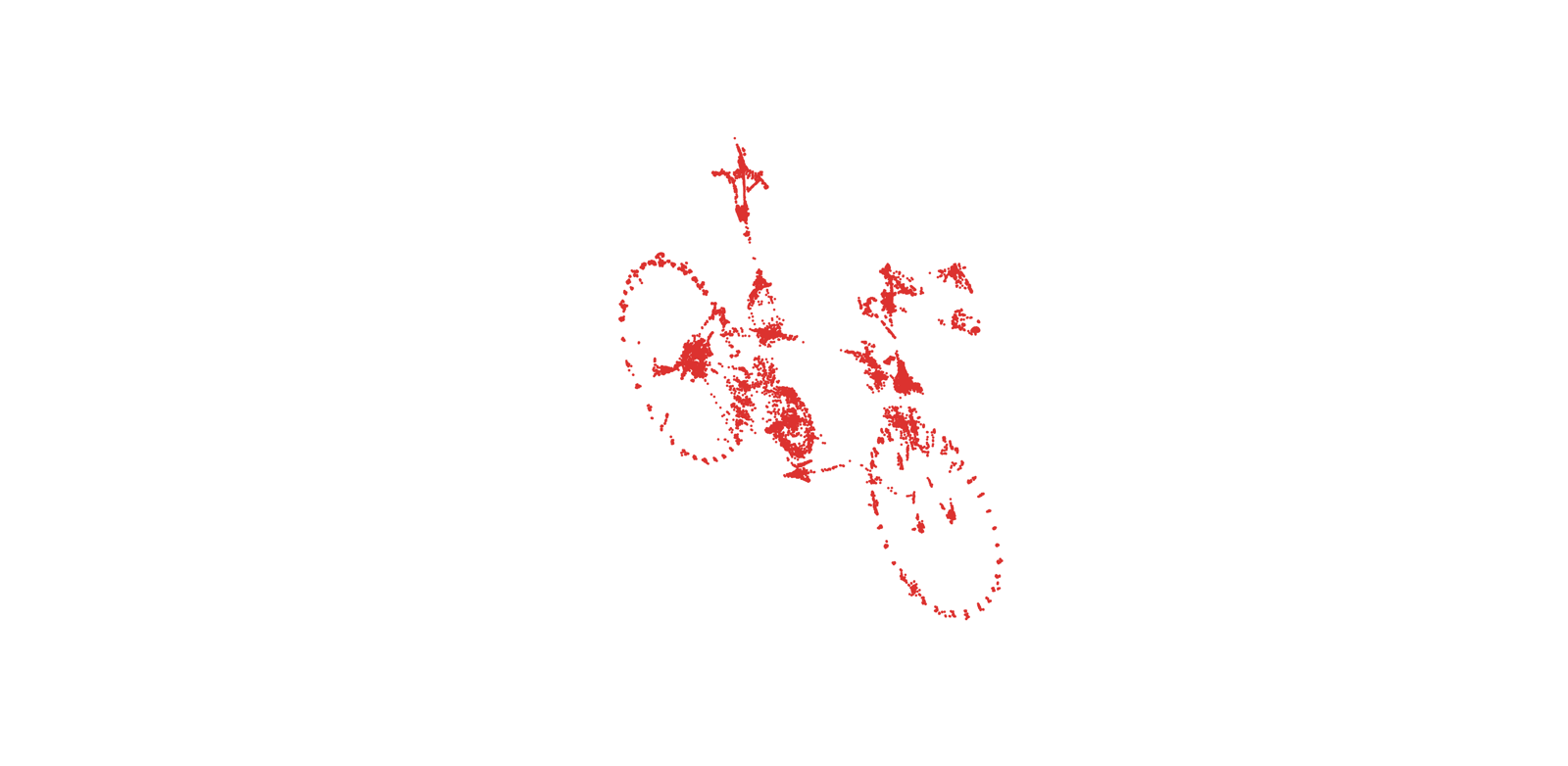}
    	\label{fig:bike_def_PC}}
   	\subfigure[$\mathcal{V'}$]{
      	\includegraphics[trim={18cm 7.5cm 16cm 7cm},clip,height=2.2cm,keepaspectratio]{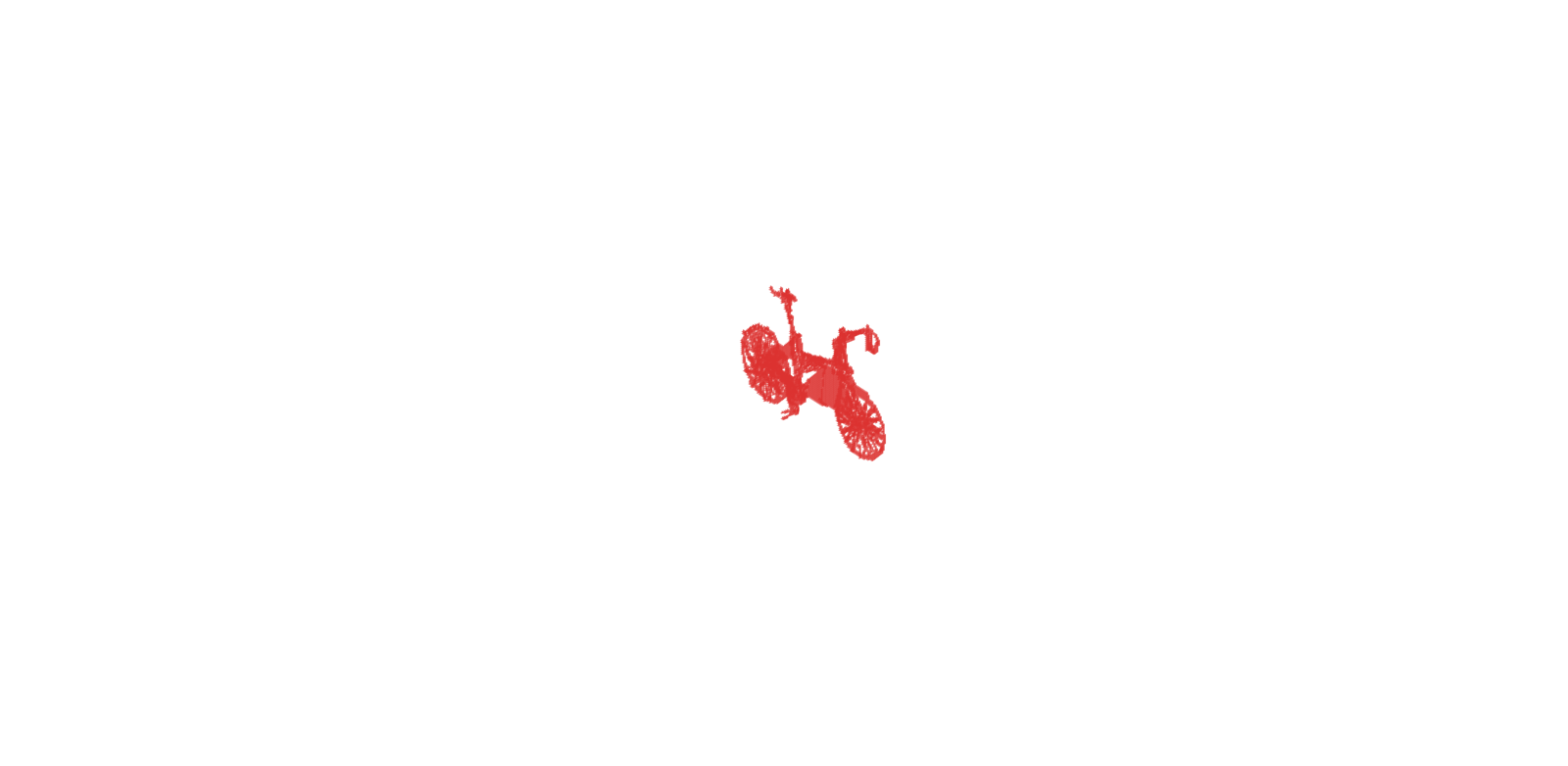}
    	\label{fig:bike_def_vxl}}
\caption{Example where the source bicycle $\mathcal{S}$ is deformed to the target $\mathcal{T}$ using our method to find dense correspondences. The point cloud and the voxel model, $\mathcal{V}_\mathcal{T}$, of $\mathcal{T}$ are shown in (c) and (d) respectively. The deformed model, $\mathcal{S'}$, and its point cloud and voxel model, $\mathcal{V'}$, are shown in (e), (f) and (g) respectively. One can realize that the point cloud of $\mathcal{S'}$ looks close to the point cloud of $\mathcal{T}$. If we only use the $s_{dist}$ metric the warped model $\mathcal{S'}$ would be considered in the graph as an edge. However, when considering its connections one can realize it looks incorrect and messy (see (e)). So if we also use the $s_{IoU}$ metric we can avoid such false positive cases.}
\label{fig:metricsExample}
\end{figure*}

\begin{table*}
\centering \small
\resizebox{0.94\textwidth}{!}{\begin{tabular}{@{}rrrrcrrrcrrrcrrr@{}} 
\toprule
& \multicolumn{3}{c}{\textbf{Kong \etal} \cite{Chen2017} \textbf{AR}} &\hphantom & \multicolumn{3}{c}{\textbf{Our \textsc{Ffd}-AR}} &\hphantom & \multicolumn{3}{c}{\textbf{Kong \etal} \cite{Chen2017} \textbf{SF}} &\hphantom & \multicolumn{3}{c}{\textbf{Our SF}} \\
\cmidrule{2-4} \cmidrule{6-8} \cmidrule{10-12} \cmidrule{14-16}
& $e_{RP}$ & $e_{pose}$ & $e_{3D}$ && $e_{RP}$ & $e_{pose}$ & $e_{3D}$ && $e_{RP}$ & $e_{pose}$ & $e_{3D}$ && $e_{RP}$ & $e_{pose}$ & $e_{3D}$ \\ \midrule

\textbf{car} 			& 45 & \textbf{0.2484} 	& 0.2541 			&& \textbf{5.1529}  & 0.3253  			& \textbf{0.1401} 	&& 40 & \textbf{0.2777} & 0.2004   			&& \textbf{9.6093}  & 0.3825  			& \textbf{0.1739}  	\\
\textbf{bicycle} 		& 32 & \textbf{0.3216} 	& 0.3424 			&& \textbf{6.3982}  & 0.4619  			& \textbf{0.3007} 	&& 41 & \textbf{0.3944} & 0.3314    		&& \textbf{9.9575}  & 0.4723  			& \textbf{0.2897}  	\\
\textbf{motorbike}		& 33 & 0.8674 			& 0.1954 			&& \textbf{6.5200}  & \textbf{0.6164}   & \textbf{0.1246} 	&& 33 & 0.8037 			& 0.1830    		&& \textbf{17.5707} & \textbf{0.6451}  	& \textbf{0.0838}  	\\
\textbf{aeroplane}		& 39 & 0.4527 			& 0.3709 			&& \textbf{5.2512}  & \textbf{0.3296}  	& \textbf{0.2488} 	&& 44 & 0.3507 			& 0.3098    		&& \textbf{14.4116} & \textbf{0.3405}  	& \textbf{0.2620}  	\\
\textbf{bus}			& 25 & \textbf{0.1699} 	& \textbf{0.0998} 	&& \textbf{4.0323}  & 0.2269  			& 0.1120 			&& 38 & 0.2148 			& 0.1197    		&& \textbf{18.9652} & \textbf{0.1807}  	& \textbf{0.0909}  	\\
\textbf{chair}			& 21 & \textbf{0.1964} 	& \textbf{0.3316} 	&& \textbf{3.7319}  & 0.2053  			& 0.3556 			&& 19 & \textbf{0.1858} & \textbf{0.3046}   && \textbf{10.1102} & 0.2165  			& 0.3087  			\\
\textbf{diningtable}	& 30 & 0.2227 			& 1.2936 			&& \textbf{4.6158}  & \textbf{0.1288}  	& \textbf{0.4485} 	&& 23 & 0.1948 			& 0.6441    		&& \textbf{12.4022} & \textbf{0.1305}  	& \textbf{0.3534} 	\\
\textbf{sofa}			& 29 & 0.3430 			& 0.4838 			&& \textbf{6.2794}  & \textbf{0.2908}  	& \textbf{0.3830} 	&& 22 & \textbf{0.2682} & 0.3872    		&& \textbf{14.0180} & 0.2926  			& \textbf{0.3461}  	\\

\bottomrule
\end{tabular}}
\caption{Results of our method and \cite{Chen2017} in terms of the 2D reprojection error, $e_{RP}$, the camera pose error, $e_{pose}$, and the 3D structure error, $e_{3D}$. We evaluate the selection of a 3D model from the graph by the anchor registration (AR) and its refinement by the silhouette fitting (SF) in eight classes.}
\label{tab:results}
\end{table*}

To build up the graph structures we used eight classes: Car, bicycle, motorbike, aeroplane, bus, chair, diningtable, and sofa. We used 30 3D models from ShapeNet with manually annotated 3D anchors. Figure~\ref{fig:nricp_plane} shows our proposed method for learning dense correspondences to build up the graph. In the first row the source aeroplane (red) is warped to fit the target (blue) by using nonrigid ICP as proposed in \cite{Chen2017}. However, one can see that if the source has a different topology from the target (\eg different wings) the nonrigid ICP is likely to fail as shown by the warped model (red). In the second row our method first brings the two aeroplanes closer to each other by deforming the source using \textsc{Ffd} to fit the target (see Figure~\ref{fig:nricp_plane_5}). One can note that the wings of the source are deformed and are moved in to close alignment to the target wings. Then, we apply nonrigid ICP between the deformed source and the target to find dense correspondences. One can see that the warped model proposed by our method is significantly more similar to the target\footnote{Videos can be found in the supplementary material.}. We used a \textsc{Ffd} lattice with $l,m,n=3$ (\ie $4 \times 4 \times 4$ grid) which gives 64 control points. Since we are performing a 3D-3D anchors registration with a few anchors we found that 64 control points are sufficient to obtain the desired deformation resolution for all object classes used in this work. Moreover, since we are imposing symmetry on the lattice we only have to optimize over half of the grid which gives us a low-dimensional parametrization.

To measure the quality of the warped model we used two metrics, the surface distance metric, $s_{dist}$, and the intersection over union (IoU) between the voxel models, $s_{IoU}$, as defined in Subsection~\ref{metrics_graph} - $s_{dist} < \theta_{dist}$ \& $s_{IoU} > \theta_{IoU}$, where $\theta_{dist}$ and $\theta_{IoU}$ are predefined thresholds. Figure~\ref{fig:metricsExample} shows an example where a source bicycle $\mathcal{S}$ is deformed to the target $\mathcal{T}$ using our method to find dense correspondences. The $s_{dist}$ between the target $\mathcal{T}$ and the warped model $\mathcal{S'}$ is of $1e^{-4}$ which is below $\theta_{dist} = 1e{-3}$. If we only use this metric the warped model $\mathcal{S'}$ would be considered in the graph as an edge. The drawback of this metric is that it relies on the vertices ignoring its connections. One can realize that the point cloud of $\mathcal{S'}$ looks close to the point cloud of $\mathcal{T}$. However, when considering its connections one can realize it looks incorrect and messy (see $\mathcal{S'}$). To handle such cases we voxelize the target $\mathcal{T}$ and the warped model $\mathcal{S'}$ to compute the IoU. The IoU between the voxel models $\mathcal{V}_\mathcal{T}$ and $\mathcal{V'}$ is $0.1694$ which is below $\theta_{IoU} = 0.25$. So the warped model is not considered as an edge in the graph since the quality is low. $\theta_{dist}$ and $\theta_{IoU}$ were defined such that we can avoid these false positive cases. By applying both metrics we can increase the quality of the graph and consequently the final 3D reconstruction. The resolution of the voxel models is of $128^3$.

\begin{figure}[b!]
	\centering
    \subfigure[Bicycle]{
       	\includegraphics[trim={12.75cm 1.5cm 11.1cm 0.5cm},clip,height=1.95cm,keepaspectratio]{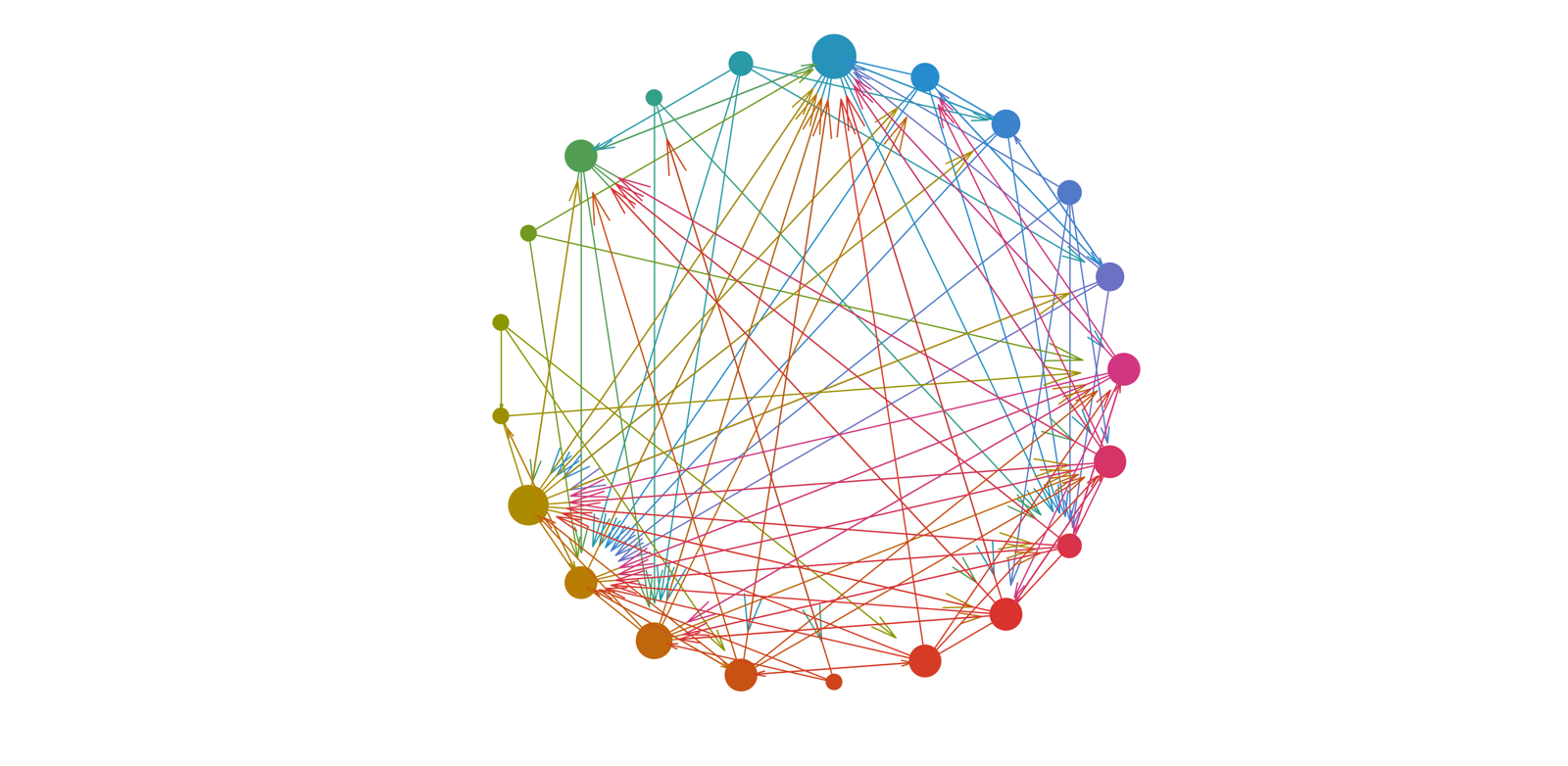}
      	\label{fig:graph_bike}}
	\subfigure[Bicycle \cite{Chen2017}]{
      	\includegraphics[trim={12.75cm 1.5cm 11.1cm 0.5cm},clip,height=1.95cm,keepaspectratio]{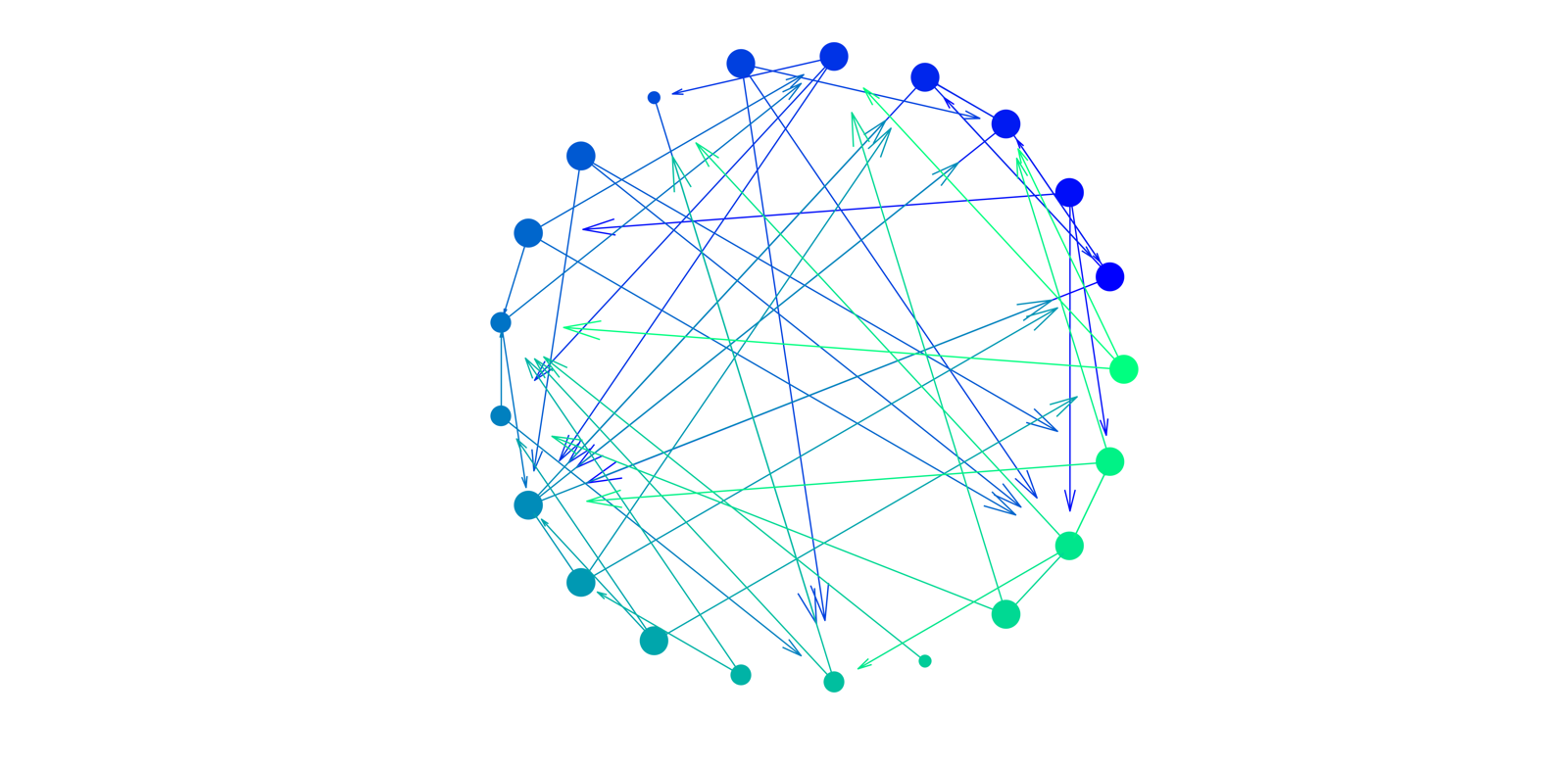}
    	\label{fig:graph_bike_ck}}
   	\subfigure[Car]{
      	\includegraphics[trim={12.55cm 1.45cm 11.1cm 0.5cm},clip,height=1.95cm,keepaspectratio]{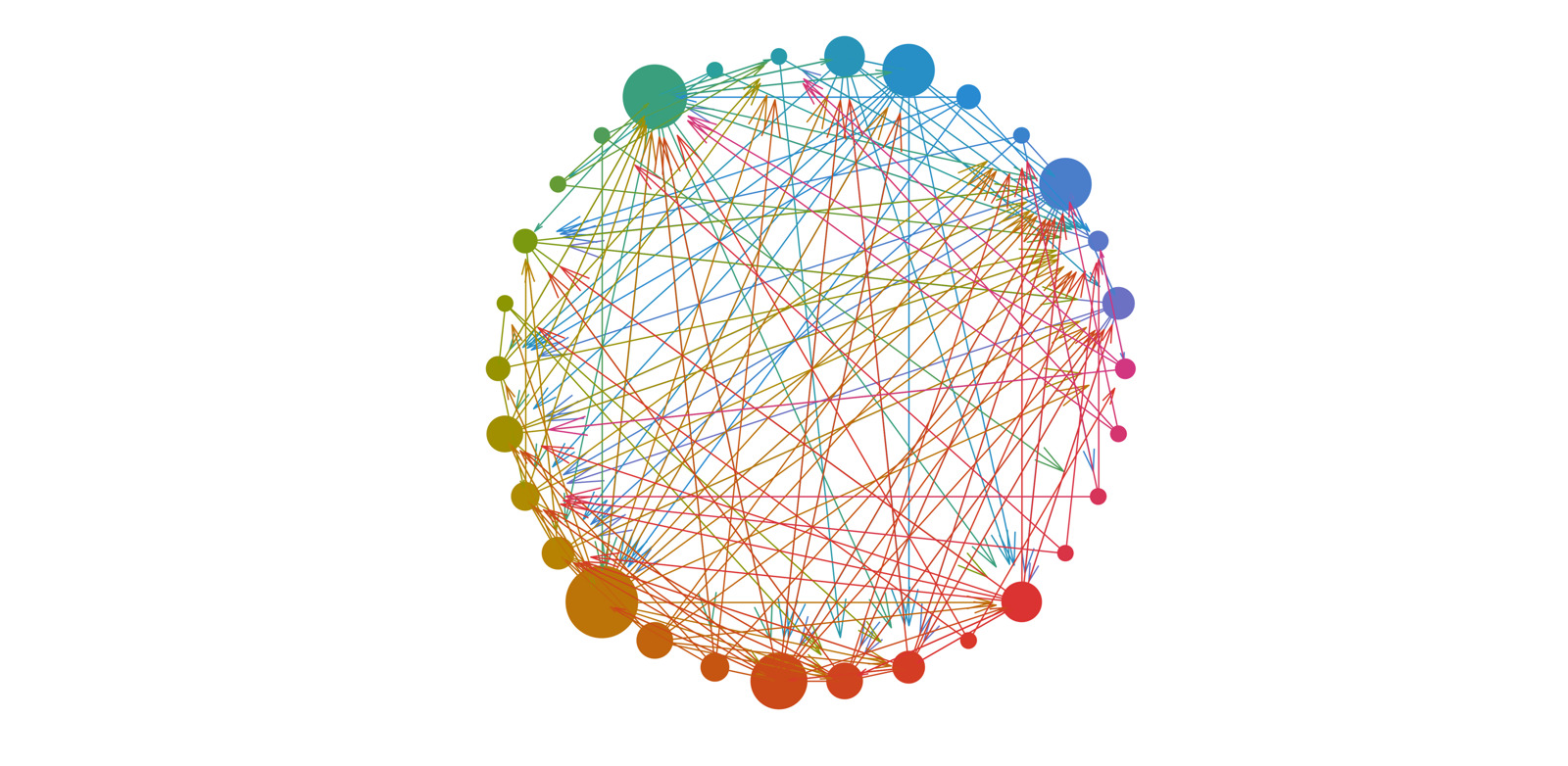}
    	\label{fig:graph_car}}
   	\subfigure[Car \cite{Chen2017}]{
      	\includegraphics[trim={12.45cm 1.5cm 11.1cm 0.5cm},clip,height=1.95cm,keepaspectratio]{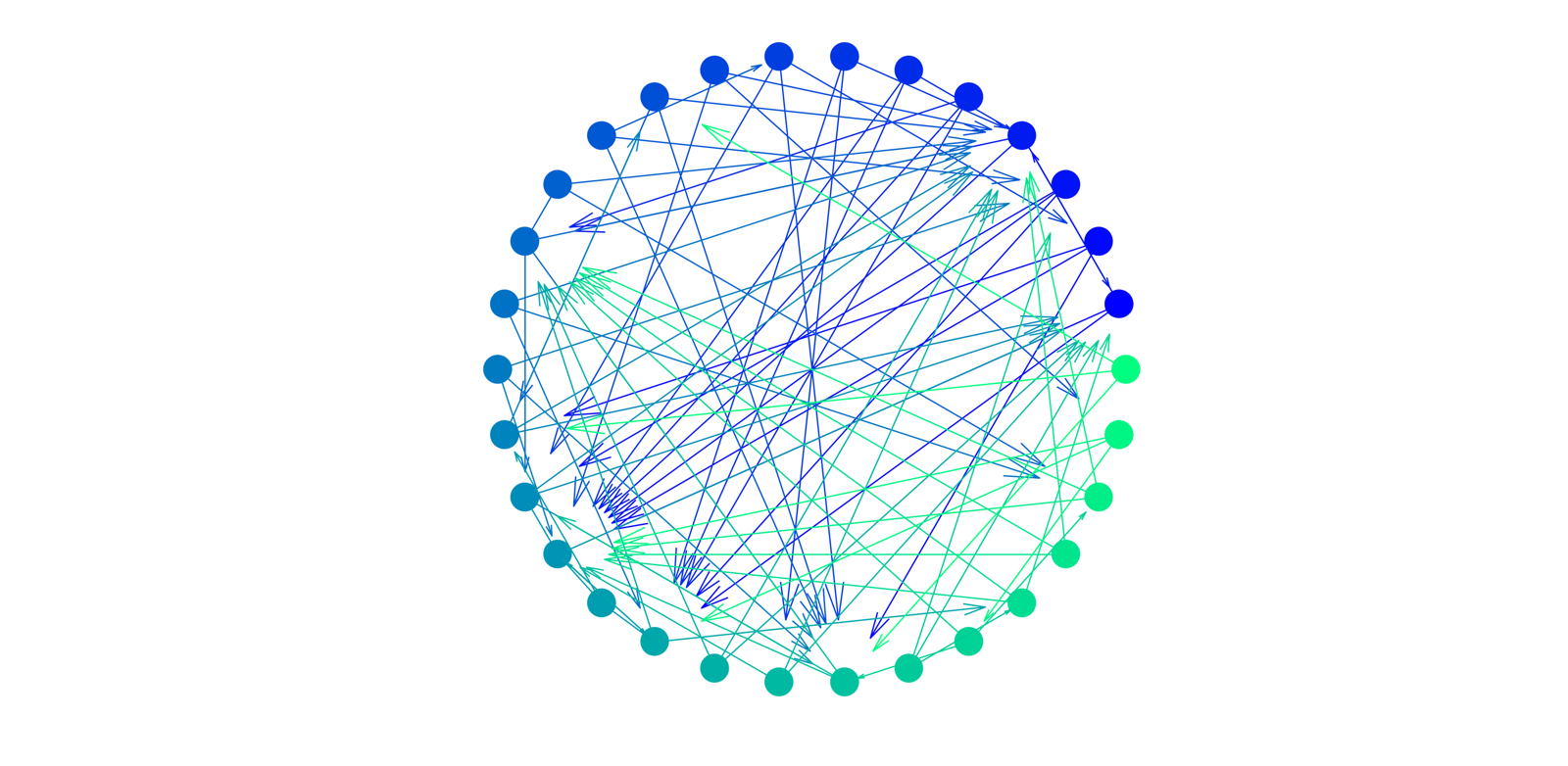}
    	\label{fig:graph_car_ck}}
\caption{Bicycle and car graphs: (a) and (c) show the graphs created by our method, and (b) and (d) show the graphs in \cite{Chen2017}.}
\label{fig:graphs}
\end{figure}

Figure~\ref{fig:graphs} shows the graph structures learned for the bicycle and car classes. Each node is a 3D model and its size illustrates its connectivity density. The size is proportional to the number of edges starting from the node. One can clearly notice that both graphs built up by our method are denser than the graphs in \cite{Chen2017} which is obvious since we have higher quality correspondences.   


\subsection{Single Image 3D Reconstruction}
We evaluate the performance of our method on PASCAL3D+ and we compared to the results presented by Kong \etal in \cite{Chen2017}. Table~\ref{tab:results} summarizes the results of our method and \cite{Chen2017} in terms of the 2D reprojection error, $e_{RP}$, the camera pose error, $e_{pose}$, and the 3D structure error, $e_{3D}$. We evaluate the selection of a 3D model from the graph by the anchor registration (AR) and its refinement by the silhouette fitting (SF). Our method clearly outperforms the method proposed in \cite{Chen2017}. Moreover, one can see that the model refinement by the SF plays an important role in the final 3D reconstruction error.

\begin{figure*}
    \centering
    \textbf{\hspace{-5pt} (a) Input \hspace{23pt} (b) \textsc{Ffd}-AR \hspace{25pt} (c) SF \hspace{25pt} (d) \textsc{Ffd}-AR \hspace{25pt} (e) SF \hspace{25pt} (f) Kong \etal \cite{Chen2017} \hspace{15pt} (g) GT}
    \includegraphics[trim={0cm 0cm 0cm 0cm},clip,width=1.01\textwidth,keepaspectratio]{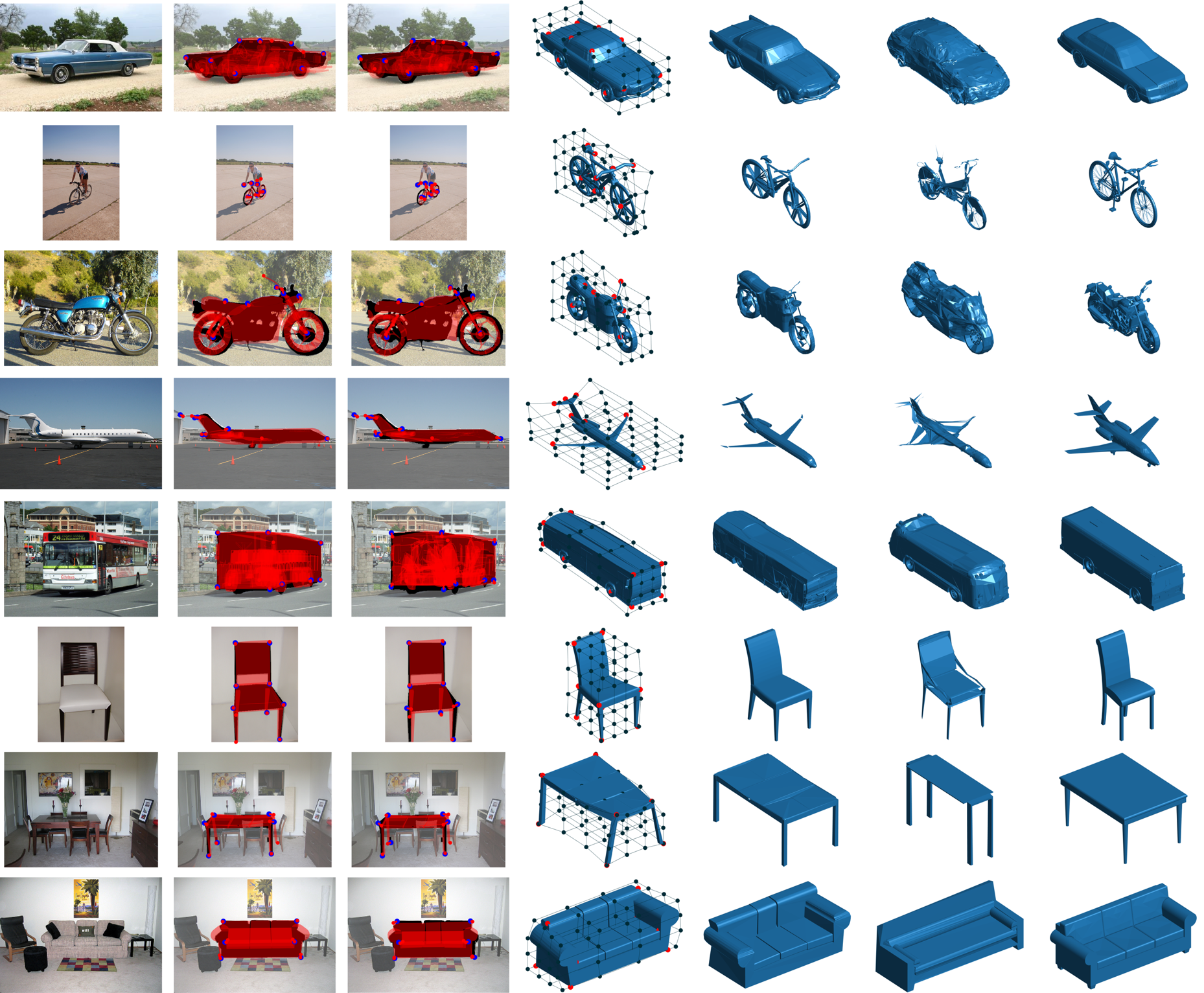}
    \caption{Qualitative results for the eight classes chosen. Column (a) shows the input natural image. Column (b) shows the \textsc{Ffd} anchor registration used for the selection of a 3D model from the graph. It shows the ground truth anchors (blue dots), the ground truth silhouette in black, the projection of the estimated anchors (red dots), and the silhouette of the estimated 3D model overlaid in red. The refinement of the 3D model by using the silhouette fitting is shown in column (c). The selected 3D model is shown in column (d) and it also displays the deformed \textsc{Ffd} lattice. The final 3D model reconstructed through the silhouette fitting is shown in column (e). We compare our results with the 3D models reconstructed in \cite{Chen2017} and the ground truth which are shown in columns (f) and (g) respectively.}
    \label{fig:bigPlot}
\end{figure*}

Qualitative results are shown in Figure~\ref{fig:bigPlot} for the eight classes chosen\footnote{Videos can be found in the supplementary material.}. Column (a) shows the input image. Column (b) shows the \textsc{Ffd} anchor registration used for the selection of a 3D model from the graph. The refinement of the 3D model by using the silhouette fitting is shown in column (c). The selected 3D model is shown in column (d). The final 3D model reconstructed through the silhouette fitting is shown in column (e). We compare our results with the 3D models reconstructed in \cite{Chen2017} and the ground truth which are shown in columns (f) and (g) respectively. One can see that our method selects a proper model and its refinement by the silhouette fitting step can be smooth. For instance, the 3D model selected for the car image in the first row is a proper choice. However, the car is still slightly deformed to better fit the silhouette. The 3D model selected for the bicycle in the second row had its handlebar grip bended down but the refinement step managed to bring it up and it looks even closer to the ground truth. One can notice that our method looks more realistic and more detailed than the method proposed by Kong \etal \cite{Chen2017}. What is interesting is that our method manages to reconstruct a 3D model from a single image even better than the ground truth which were manually selected by humans. Car, motorbike and aeroplane are good examples for showing that. Humans are not good in choosing 3D models to fit an image. Our method can take care of details that we humans can easily ignore.

\section{Conclusion}
We demonstrated that a low-dimensional \textsc{Ffd} parametrization and sparse linear representation are able to compactly model the intrinsics deformation across a class of 3D CAD models. We showed that dense 3D reconstruction from a single image can be performed using our compact model representation given the 2D anchors and the object silhouette. Experiments revealed that our approach is able to reconstruct detailed and realistic 3D meshs which extends the applicability of our method.



{\small
\bibliographystyle{ieee}
\bibliography{egbib}
}

\end{document}